\documentclass{article}
\pdfoutput=1

\usepackage{microtype}
\usepackage{graphicx}
\usepackage{subcaption}
\usepackage{booktabs} 

\usepackage{hyperref}

\usepackage[accepted]{arXiv}

\usepackage{amsmath}
\usepackage{amssymb}
\usepackage{mathtools}
\usepackage{amsthm}

\usepackage{amsmath,amsfonts,bm}

\def\eqref#1{equation~\ref{#1}}

\def\1{\bm{1}}

\DeclareMathAlphabet{\mathsfit}{\encodingdefault}{\sfdefault}{m}{sl}
\SetMathAlphabet{\mathsfit}{bold}{\encodingdefault}{\sfdefault}{bx}{n}

\usepackage{hyperref}
\usepackage{url}
\usepackage{graphicx}
\usepackage{booktabs}
\usepackage{subcaption}
\usepackage{pgfplots}
\usepackage{array}
\usepackage{booktabs}
\usepackage{multirow}
\usepackage{enumitem}
\usepackage{xcolor}
\usepgfplotslibrary{dateplot}
\usepgfplotslibrary{groupplots}
\usepgfplotslibrary{fillbetween}
\usepgfplotslibrary{statistics}
\pgfplotsset{
        compat=1.3,
        blackboxplot/.style={
            boxplot,
            draw=black,
            solid,
            fill=white,
            mark=*,
            every mark/.append style={
                fill=black,
                mark size=0.3pt
            },
        },
    }

\usepackage[capitalize,noabbrev]{cleveref}

\theoremstyle{plain}

\theoremstyle{definition}

\theoremstyle{remark}

\usepackage[textsize=tiny]{todonotes}

\icmltitlerunning{Loci-Segmented: Improving Scene Segmentation Learning}
\usepackage{stfloats}
\usepackage{afterpage}

\begin{document}

\twocolumn[
\icmltitle{Loci-Segmented: \\
Improving Scene Segmentation Learning}




\begin{icmlauthorlist}
\icmlauthor{Manuel Traub}{yyy}
\icmlauthor{Frederic Becker}{yyy}
\icmlauthor{Adrian Sauter}{yyy}
\icmlauthor{Sebastian Otte}{yyy}
\icmlauthor{Martin V. Butz}{yyy}
\end{icmlauthorlist}

\icmlaffiliation{yyy}{Neuro-Cognitive Modeling Group, University of Tübingen, Germany}

\icmlcorrespondingauthor{Manuel Traub}{manuel.traub@uni-tuebingen.de}
\icmlcorrespondingauthor{Frederic Becker}{frederic.becker@uni-tuebingen.de}

\icmlkeywords{object segmentation, object tracking, compositional representation learning, neuro-cognitive modeling, machine Learning, ICML}

\vskip 0.3in
]



\printAffiliationsAndNotice{}  

\def\locis*{Loci-s}
\def\locisd*{Loci-$s_d$}
\def\locil*{Loci-l}
\def\loci*{Loci}
\begin{figure*}[b]
    \centering
    \setlength{\fboxsep}{0pt}
    \setlength{\fboxrule}{1pt}
    
    \newlength{\mysubfloatwidthfive}
    \setlength{\mysubfloatwidthfive}{\dimexpr(\linewidth)/7\relax}

    \begin{subfigure}[b]{\mysubfloatwidthfive}
        \fbox{\includegraphics[width=\textwidth]{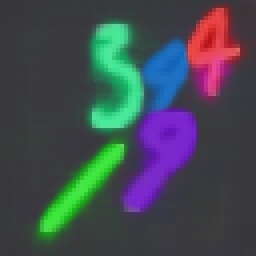}}
    \end{subfigure}
    \hfill 
    \begin{subfigure}[b]{\mysubfloatwidthfive}
        \fbox{\includegraphics[width=\textwidth]{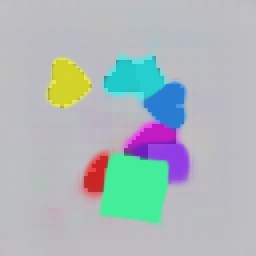}}
    \end{subfigure}
    \hfill 
    \begin{subfigure}[b]{\mysubfloatwidthfive}
        \fbox{\includegraphics[width=\textwidth]{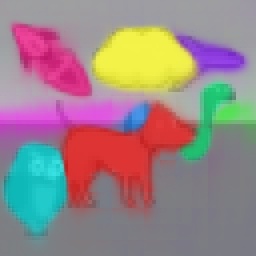}}
    \end{subfigure}
    \hfill 
    \begin{subfigure}[b]{\mysubfloatwidthfive}
        \fbox{\includegraphics[width=\textwidth]{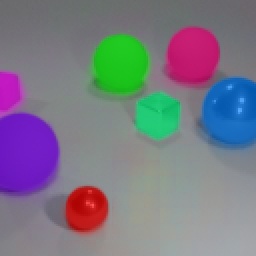}}
    \end{subfigure}
    \hfill 
    \begin{subfigure}[b]{\mysubfloatwidthfive}
        \fbox{\includegraphics[width=\textwidth]{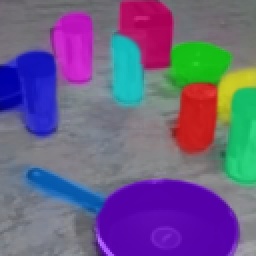}}
    \end{subfigure}  
    \hfill 
    \begin{subfigure}[b]{\mysubfloatwidthfive}
        \fbox{\includegraphics[width=\textwidth]{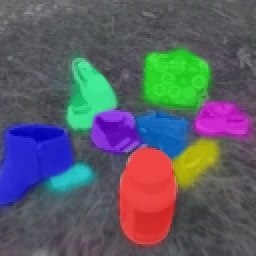}}
    \end{subfigure}  
    
    \caption{Exemplar slot-based autoregressive segmentations inferred by \locis*, generalizing to 7-10 objects while being trained on only 4-6 objects. Shown are moving MNIST digits and dSprites, the Abstract Scene dataset, CLEVR, SHOP VRB, and a combination of GSO and HDRI-Haven, all of which were considered in a recent compositional scene understanding review \cite{Yuan:2023}.}
    \label{fig:occlusion}
\end{figure*}

\begin{abstract}
Current slot-oriented approaches for compositional scene segmentation from images and videos rely on provided background information or slot assignments.
We present a segmented location and identity tracking system, Loci-Segmented (\locis*), which does not require either of this information.
It learns to dynamically segment scenes into interpretable background and slot-based object encodings, separating rgb, mask, location, and depth information for each.
The results reveal largely superior video decomposition performance in the MOVi datasets and in another established dataset collection targeting scene segmentation.
The system's well-interpretable, compositional latent encodings may serve as a foundation model for downstream tasks.
\end{abstract}

\section{Introduction}


Visual scene understanding from images or videos presents unique challenges. Classical architectures such as CNNs \citep{liu2022convnet} or ViTs (Vision Transformers) \citep{dosovitskiy2020image} exacerbate existing limitations, being data-hungry, susceptible to adversarial attacks, and low on interpretability. 
To address these challenges, slot attention mechanisms have emerged as a promising avenue \citep{locatello2020object}.
These architectures address critical challenges associated to the binding problem \cite{Greff:2020}. 
In particular, they offer a way to bind features into 'slots' that dynamically represent distinct entities in a scene \citep{locatello2020object}, building upon prior work in attention mechanisms \citep{vaswani2017attention} and capsule networks \citep{sabour2017dynamic}.
Current state-of-the-art systems include the highly potent Slot Attention for Video model (SAVi++, \citealp{Elsayed:2022}) and the location and identity tracking slot-based recurrent architecture (\loci*, \citealp{Traub:2023}). Still, both systems have a weakness: 
SAVi++ relies on supervised slot assignments upon trial initialization while \loci* relies on the provision of static background information. 

\loci* has shown superior performance on the CATER challenge, in which objects (balls and cones) are transported hidden within other objects (cones).
It is rather closely related to other slot-based object processing architectures including SAVi++ \citep{Elsayed:2022,locatello2020object,Kipf:2022,Wu:2023slotformer}, surveyed in \citep{Yuan:2023}.
It differs in (i) its slot-specific encoding approach that starts from pixels, (ii) its emergent disentanglement of objects from positions, and (iii) its event-oriented internal processing loop. 
However, its reliance on a provided, static background module prevented processing more complex dynamic backgrounds or moving cameras. 
Moreover, \loci* was not able to profit from or predict depth information.

In this work, we enable \loci* to deal with (i) dynamic complex backgrounds, (ii) videos where the camera is moving, and (iii) depth information without initial slot assignments and without the provision of background information during evaluation.  
Additional improvements enable the segmentation of scenes with more complex and diverse objects.
Thereby, we enhance the state-of-the-art of scene segmentation algorithms in both the MOVi-* datasets \citep{greff2022} as well as in a scene segmentation dataset benchmark suite used in a recent review paper \citep{Yuan:2023}.
Our key contributions are:

\begin{enumerate}[label=\textbf{(\roman*)}]
\item We introduce Loci-Segmented a compositional slot-based video segmentation model that makes several key improvements over related slot-based compositional scene segmentation algorithms \citep{Traub:2023,Yuan:2023,greff2022} by introducing a \textit{background module}, implementing enhanced \textit{encoder} and \textit{decoder} pipelines, and including \textit{Scene-Relative-Depth} as target and, optional, as input.
\item Demonstrating superior segmentation performance on the challenging multi-object video benchmark (MOVi) \cite{greff2022} through extensive supervised pertaining while not relying on ground truth slot initialization during testing like previous approaches.
\item Generalization to and largely outperforming all other methods on a recently introduced compositional scene segmentation benchmark suite \citep{Yuan:2023}.
\item Providing well interpretable latent codes that by design disentangle mask, depth and texture (rgb) codes. 
\end{enumerate}

\begin{figure*}[t]
    \centering
    \includegraphics[width=\linewidth]{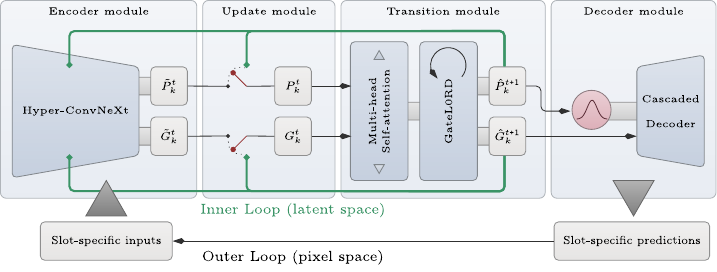}
    \caption{The primary \locis* architecture features: a Hyper-ConvNeXt encoder, which generates Position and Gestalt codes slot-individually; an Update module, which adaptively fuses current encoder information with prior temporal predictions; a Transition module, which calculates object dynamics via a GateL0RD layer (a strongly gated RNN, cf. \citealp{Gumbsch:2021c}) and inter-slot interactions via self-attention; 
    finally, a Decoder module
    computing sequential slot-wise predictions including depth estimates for the subsequent frame.}
    \label{fig:loci}
\end{figure*}


\section{Loci Architecture}

Before detailing the novel extensions in \locis*, we briefly introduce \loci* \citep{Traub:2023}, a slot-based object-oriented processing architecture that consists of a slot-wise encoder, a transition, and a decoder module (cf., \autoref{fig:loci}).

{\bf Encoder Module:}
In contrast to other slot-based approaches \citep{Yuan:2023}, \loci* slots each start from the input image. At time point $t$, each slot $k$ receives as input 
the actual video frame $I^t$, the previous prediction error $E^t$, and a background mask $\hat{M}^t_{bg}$.
Moreover, to focus each slot on its own object encoding, previous slot predictions are fed in as additional input, including its predicted position $\hat{Q}^t_k$ encoded as an isotropic Gaussian in pixel space, its visibility mask $\hat{M}^{t,v}_k$ and object mask $\hat{M}^{t,o}_k$ encoded as grayscale images, its RGB image $\hat{R}^t_k$, and the summed visibility masks of the remaining slots $\hat{M}^{t,s}_k$.

As output, the encoder produces two types of codes for each slot: 
{\bf Gestalt Codes $\tilde{G}^t_k$:} A 1D latent representations of an object's appearance, capturing shape, color, texture, and other visual attributes;
{\bf Position Codes $\tilde{P}^t_k$:} A spatial code including the object's 2D location ($x_k,y_k$), its size ($\sigma_k$), and its distance in depth encoded as a priority code ($\rho_k$).

{\bf Transition Module: }
The transition module contains a slot-wise recurrent module and a multi-head attention module. 
The recurrent module implements GateL0RD units, which encode LSTM-like recurrent cells with an even stronger shielding, to foster event-predictive encodings \citep{Gumbsch:2021c}. 

The multi-head attention module enables the object-interaction-oriented exchange of information between slots. 
As its result, the transition module outputs next object-respective Gestalt-Codes $\hat{G}^{t+1}_k$ and positions $\hat{P}^{t+1}_k$. 

{\bf Decoder Module:}
The decoder reconstructs the predicted scene starting from a 3D tensor that combines the Gestalt code as channels with the positional encoding ($x_k,y_k$,$\sigma_k$).

\newpage
It then upscales this tensor to the full input resolution via a ResNet. 
The outputs are an RGB slot image $\hat{R}^{t+1}_k$, visibility mask $\hat{M}^{t+1,v}_k$, and position $\hat{Q}^{t+1}_k$. 
The scene is finally recomposed by combining the masked RGB outputs with respect to their respective priority codes $\hat{\rho_k}$ and the background mask. 

\section{Methodology}

\locis* builds on \loci* but significantly enhances its abilities: We enable the processing and prediction of depth information; we design an even more dynamic encoder-decoder framework; and we introduce a dedicated background processing module. Detailed \locis* network wiring and size information can be found in \autoref{app:lociswiring}.

\subsection{Depth as Input}
We introduce a novel input channel to the \locis* model, denoted as Scene-Relative Depth (see \autoref{app:depth_norm} for more details).
Depth normalization is expected to significantly support object segmentation, as object edges will naturally be marked by spatial depth non-linearities. 

\begin{figure}[t]
    \centering
    \includegraphics[width=\linewidth]{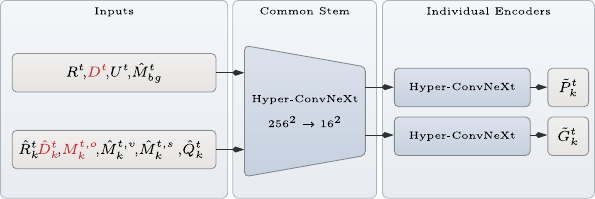}
    \caption{New Encoder design with the added depth information and slot-wise depth and object mask channels (in red). One encoder head processes both common (RGB frame $R^t$, depth frame $D^t$, uncertainty mask $U^t$, background mask $\hat M^t_{bg}$) and slot specific inputs (decoder outputs from the previous iteration: slot rgb $\hat R^t_k$, slot depth $\hat D^t_k$, amodal mask $M^{t,o}_k$, visibility mask $\hat M^{t,v}_k$, summed masks from other slots $\hat M^{t,s}_k$, the 2d Gaussian position $\hat Q^t_k$). In addition to the top-down feedback provided by the slot specific inputs, Hyper-ConvNeXt blocks also provide top-down feedback in the form of dynamic weight residuals computed from predicted Gestalt-Codes $\hat G^t_k$.}
    \label{fig:encoder}
\end{figure}
\begin{figure}[t]
    \includegraphics[width=\linewidth]{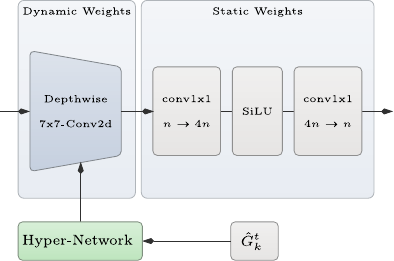}
    \caption{A single Hyper-ConvNeXt block within the encoder where a top-down hyper-network translates the Gestalt-Code predicted in the last iteration $\hat{G_k^{t}}$ into spatial convolutional kernel weight residuals augmenting the receptive field of the particular slot encoder to be more susceptible to the previously encoded entity.}
    \label{fig:hyperconvnext}
\end{figure}

\subsection{Encoder}
The encoder and decoder subnetworks adopt a ConvNeXt-like architecture \cite{liu2022convnet}, replacing the previously used ResNet architecture.
Furthermore, we add an inner top-down processing loop to the architecture, which propagates
predicted Gestalt code information $\hat{G_k^{t}}$ directly into the encoder. These codes are utilized within a hypernetwork to compute dynamic residuals for the depth-wise convolutional kernels present in the ConvNeXt blocks of the encoder (see \autoref{fig:encoder}).
This architectural modification enables the encoder to integrate top-down feedback into its computations, thereby improving the object-specific encoding pipeline.

\subsection{Decoder}
\begin{figure}[t]
    \centering
    \includegraphics[width=\linewidth]{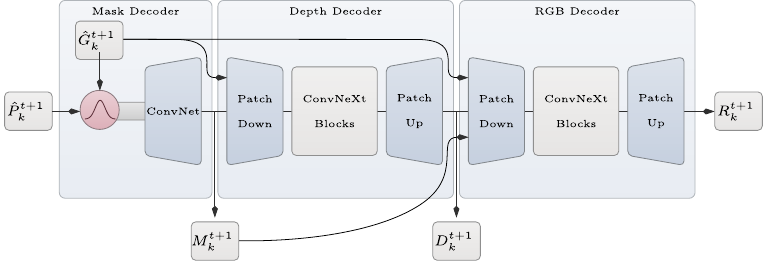}
    \caption{Decoder visualization illustrating the cascaded reconstruction strategy, first decoding the mask, then the depth, and finally the RGB image of a slot-encoded entity. Specifically the Gestalt code is partitioned into three equi-length segments of 256 elements each. The Mask Decoder is implemented by a compact convolutional network; its input comprises the Mask-Gestalt Code $\hat{Gm^{t}_k}$modulated by a 2D Gaussian heatmap, which is derived from the Position Code. The Depth Decoder features a U-Net architecture with aggressive down-sampling and up-sampling pathways, altering the spatial resolution by a factor of 16. This Depth Decoder receives as input the Depth-Gestalt Code $\hat{Gd^{t}_k}$ modulated by the mask output from the Mask Decoder. The RGB Decoder operates on the same principle as the Depth Decoder but incorporates an additional input: the depth map generated by the Depth Decoder.
    }
    \label{fig:decoder}
\end{figure}

We introduce a cascaded decoder architecture, shown in \autoref{fig:decoder}, and partition the Gestalt Code $G^{t}_k$ into three segments, each comprising 256 elements. These segments encode mask $Gm^{t}_k$, depth $Gd^{t}_k$, and RGB channels $Gr^{t}_k$. 
The Mask Decoder module uses element-wise multiplication between the Gestalt Code $\hat{Gm^{t}_k}$ and a two-dimensional isotropic Gaussian heatmap generated from Position Code $\hat{P^{t}_k}$. This modulated spatialized Gestalt Code is then subject to a compact convolutional neural network.
The Depth Decoder module is implemented by a U-Net architecture. It decodes the depth information via the predicted Depth Gestalt Code $\hat{Gd^{t}_k}$, which is multiplied layer-wise with the computed object mask, enforcing masked outlines. 
The RGB Decoder module clones the Depth Decoder architecture but additionally receives the Depth Decoder's output as input. 
Finally, the RGB image of the encoded object is reconstructed via the RGB Gestalt code, which is layer-wise multiplied with the computed object mask and additionally informed by the generated depth estimations. 

The cascading of the decoder architecture does not only encourage a disentangled encoding of an object's mask (i.e., its shape), its distance to the camera (i.e., its depth), and its appearance, but it also facilitates appearance reconstruction because the prediction of the mask is easier and then informs the depth and RGB-pattern reconstruction. Additionally, the cascaded decoder facilitates the reconstruction of the unoccluded raw mask $\hat M_k^{t,o}$ and the occlusion-aware mask $\hat M_k^t$, because only the Mask-decoder, but not the Depth and RGB decoders, needs to be re-run to generated the occlusion-aware mask.

\subsection{Background}
\begin{figure}[t]
    \centering
    \includegraphics[width=\linewidth]{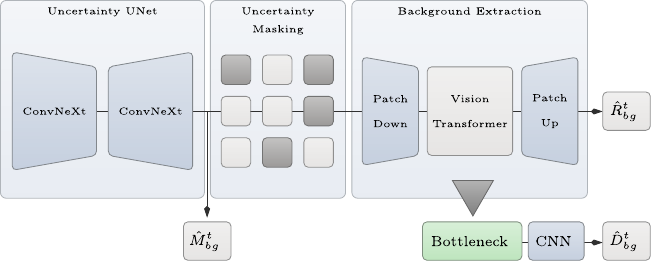}
    \caption{Background module: Input RGB or RGB+Depth is used to compute an Uncertainty or Foreground Mask via a U-Net. The mask is used in a Masked Autoencoder to reconstruct a background representation from the foreground masked input image (RGB or RGB+Depth). Within this process, input patches characterized by high uncertainty are effectively masked out. Moreover, the Depth background reconstruction is subjected to a bottleneck layer, which encourages to encode a regular background without any non-liner foreground depth masks, thereby increasing the complexity for the autoencoder to accurately reconstruct foreground objects. 
    }
    \label{fig:background}
\end{figure}

Another pivotal enhancement in our work is the development of a Background Module, which is trained prior to the slotted architectural components. This module enables the application of \locis* to environments with complex backgrounds, featuring both intricate backgrounds and moving cameras. As delineated in \autoref{fig:background}, this module is bifurcated into two core elements: an Uncertainty Network and a Background Extraction Network.

The Uncertainty Network employs a U-Net architecture with ConvNeXt residual blocks. Skip connections between down-sampling and up-sampling layers avoid vanishing gradients. 
The network is trained in a supervised manner to compositionally segment the foreground in a scene, generating an uncertainty mask that predicts the provided foreground mask from either pure RGB or RGB+Depth depending on the used version (\locis* or \locisd*).
It thus learns to deem dynamic foreground objects `uncertain', in contrast with the generally stable background elements in natural scenes. 

The output from the Uncertainty Network serves as a masking function for the Background Extraction Network. This network implements a masked autoencoder using a Vision Transformer. This Vision Transformer is designed to predict both the RGB values and the depth map from either RGB alone or from both RGB and depth maps . By selectively masking-out foreground objects, we introduce a bias favoring the exclusive reconstruction of the background elements. To further accentuate this bias, we constrain the depth reconstruction module with a narrow bottleneck. The network is trained as a background autoencoder 
by masking the reconstruction loss with the inverse of the foreground mask.

\subsection{Pretraining object encodings and decodings}
The original \loci* architecture was fully trained end-to-end without any information on objects whatsoever. 
While \locis* could also be trained in this way, in order to speed-up learning and save computational resources, we implement a sequential supervised pre-training strategy specifically tailored for \locis*'s encoder and decoder components. It is trained on single-object detection and reconstruction tasks. 

To initialize it, the encoder is feed with an input frame with all slot-specific inputs nullified except for the slot-specific 2D Gaussian position, which is computed from the ground-truth target mask during this training stage. To avoid reliance on exact position encodings, the encoding is subjected to stochastic perturbations while ensuring its confinement within the mask's boundary.
Any slot-specific inputs dependent on other slots are explicitly nullified.
Note that during this pretraining stage we thus encourage slots to encode particular masks, but we do not initialize the slots explicitly to ground-truth bounding boxes.
During all evaluations, we do not provide any supervised slot information whatsoever. 
This is in stark contrast to SAVi++, where slots are initialized to the ground-truth bounding boxes in the first frame during training and evaluation \citep{Elsayed:2022}.

 \subsection{Segmentation Preprocessing}
\begin{figure*}[t]
    \centering
    \setlength{\fboxsep}{0pt}
    \setlength{\fboxrule}{1pt}
    
    \newlength{\mysubfloatwidthfour}
    \setlength{\mysubfloatwidthfour}{\dimexpr(\linewidth)/6\relax}

    \begin{subfigure}[b]{\mysubfloatwidthfour}
        \fbox{\includegraphics[width=\textwidth]{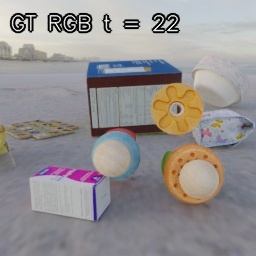}}
    \end{subfigure}
    \hfill 
    \begin{subfigure}[b]{\mysubfloatwidthfour}
        \fbox{\includegraphics[width=\textwidth]{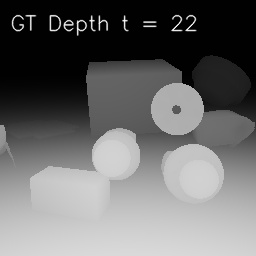}}
    \end{subfigure}
    \hfill 
    \begin{subfigure}[b]{\mysubfloatwidthfour}
        \fbox{\includegraphics[width=\textwidth]{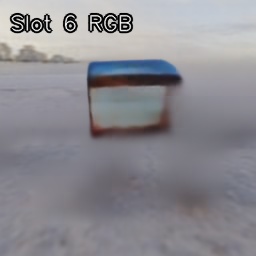}}
    \end{subfigure}
    \hfill 
    \begin{subfigure}[b]{\mysubfloatwidthfour}
        \fbox{\includegraphics[width=\textwidth]{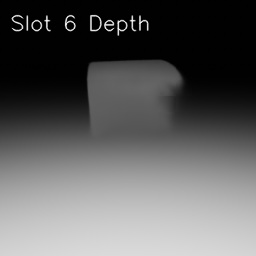}}
    \end{subfigure}
    \hfill 
    \begin{subfigure}[b]{\mysubfloatwidthfour}
        \fbox{\includegraphics[width=\textwidth]{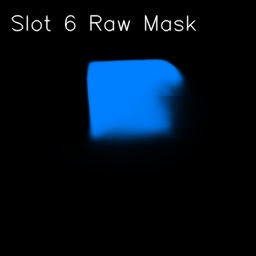}}
    \end{subfigure}    
    \\

    \vspace{10pt}
    \centering
    \setlength{\fboxsep}{0pt}
    \setlength{\fboxrule}{1pt}
    
    \newlength{\mysubfloatwidththree}
    \setlength{\mysubfloatwidththree}{\dimexpr(\linewidth)/11\relax}

    \begin{subfigure}[b]{\mysubfloatwidththree}
        \fbox{\includegraphics[width=\textwidth]{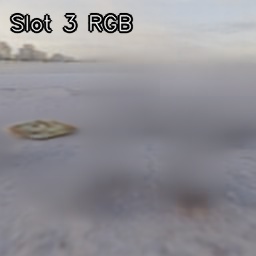}}
    \end{subfigure}
    \hfill 
    \begin{subfigure}[b]{\mysubfloatwidththree}
        \fbox{\includegraphics[width=\textwidth]{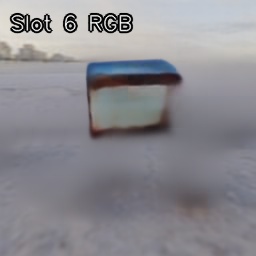}}
    \end{subfigure}
    \hfill 
    \begin{subfigure}[b]{\mysubfloatwidththree}
        \fbox{\includegraphics[width=\textwidth]{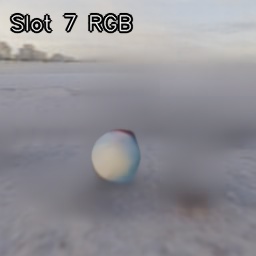}}
    \end{subfigure}
    \hfill 
    \begin{subfigure}[b]{\mysubfloatwidththree}
        \fbox{\includegraphics[width=\textwidth]{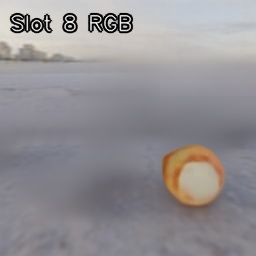}}
    \end{subfigure}
    \hfill 
    \begin{subfigure}[b]{\mysubfloatwidththree}
        \fbox{\includegraphics[width=\textwidth]{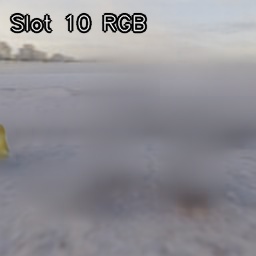}}
    \end{subfigure}
    \hfill 
    \begin{subfigure}[b]{\mysubfloatwidththree}
        \fbox{\includegraphics[width=\textwidth]{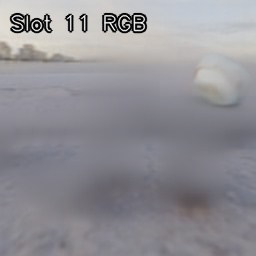}}
    \end{subfigure}
    \hfill 
    \begin{subfigure}[b]{\mysubfloatwidththree}
        \fbox{\includegraphics[width=\textwidth]{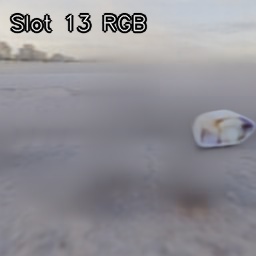}}
    \end{subfigure}
    \hfill 
    \begin{subfigure}[b]{\mysubfloatwidththree}
        \fbox{\includegraphics[width=\textwidth]{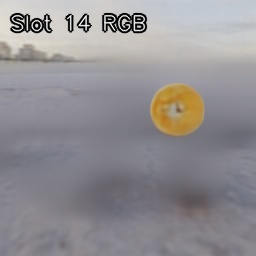}}
    \end{subfigure}
    \hfill 
    \begin{subfigure}[b]{\mysubfloatwidththree}
        \fbox{\includegraphics[width=\textwidth]{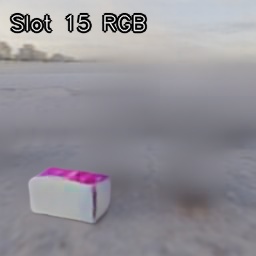}}
    \end{subfigure}
    \hfill 
    \begin{subfigure}[b]{\mysubfloatwidththree}
        \fbox{\includegraphics[width=\textwidth]{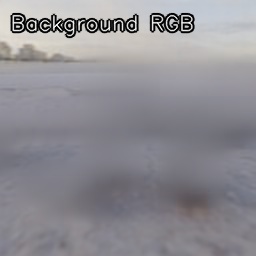}}
    \end{subfigure}
    \\

    \begin{subfigure}[b]{\mysubfloatwidththree}
        \fbox{\includegraphics[width=\textwidth]{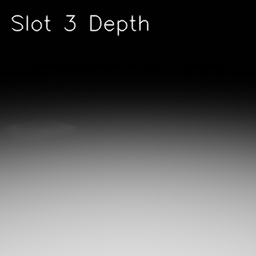}}
    \end{subfigure}
    \hfill 
    \begin{subfigure}[b]{\mysubfloatwidththree}
        \fbox{\includegraphics[width=\textwidth]{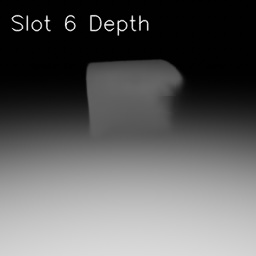}}
    \end{subfigure}
    \hfill 
    \begin{subfigure}[b]{\mysubfloatwidththree}
        \fbox{\includegraphics[width=\textwidth]{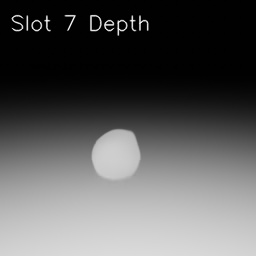}}
    \end{subfigure}
    \hfill 
    \begin{subfigure}[b]{\mysubfloatwidththree}
        \fbox{\includegraphics[width=\textwidth]{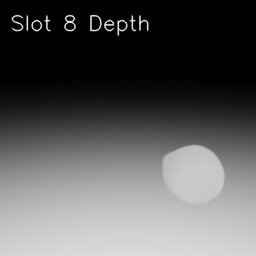}}
    \end{subfigure}
    \hfill 
    \begin{subfigure}[b]{\mysubfloatwidththree}
        \fbox{\includegraphics[width=\textwidth]{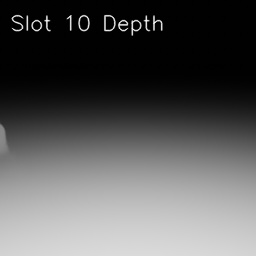}}
    \end{subfigure}
    \hfill 
    \begin{subfigure}[b]{\mysubfloatwidththree}
        \fbox{\includegraphics[width=\textwidth]{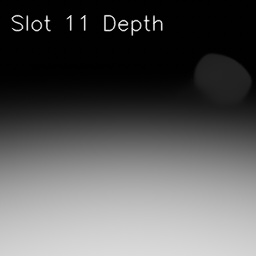}}
    \end{subfigure}
    \hfill 
    \begin{subfigure}[b]{\mysubfloatwidththree}
        \fbox{\includegraphics[width=\textwidth]{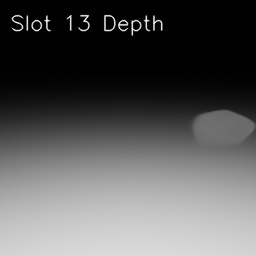}}
    \end{subfigure}
    \hfill 
    \begin{subfigure}[b]{\mysubfloatwidththree}
        \fbox{\includegraphics[width=\textwidth]{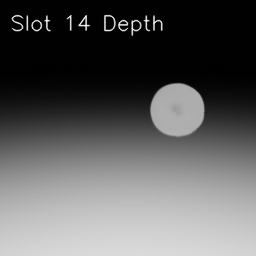}}
    \end{subfigure}
    \hfill 
    \begin{subfigure}[b]{\mysubfloatwidththree}
        \fbox{\includegraphics[width=\textwidth]{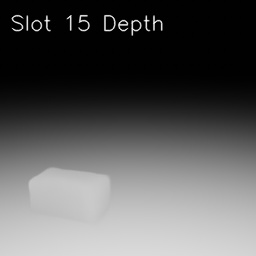}}
    \end{subfigure}
    \hfill 
    \begin{subfigure}[b]{\mysubfloatwidththree}
        \fbox{\includegraphics[width=\textwidth]{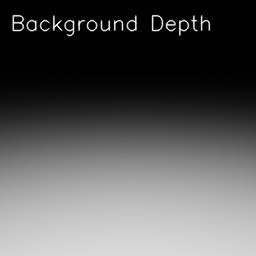}}
    \end{subfigure}
    \\
    
    \begin{subfigure}[b]{\mysubfloatwidththree}
        \fbox{\includegraphics[width=\textwidth]{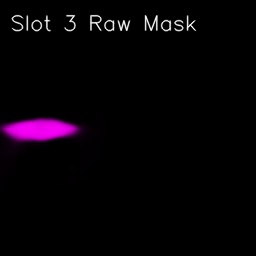}}
    \end{subfigure}
    \hfill 
    \begin{subfigure}[b]{\mysubfloatwidththree}
        \fbox{\includegraphics[width=\textwidth]{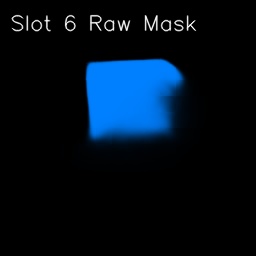}}
    \end{subfigure}
    \hfill 
    \begin{subfigure}[b]{\mysubfloatwidththree}
        \fbox{\includegraphics[width=\textwidth]{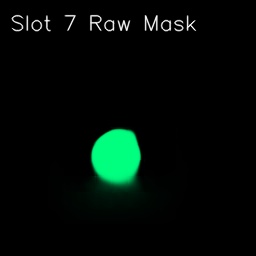}}
    \end{subfigure}
    \hfill 
    \begin{subfigure}[b]{\mysubfloatwidththree}
        \fbox{\includegraphics[width=\textwidth]{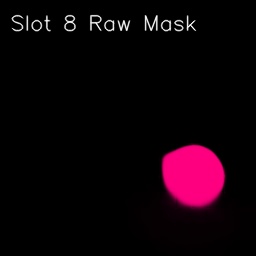}}
    \end{subfigure}
    \hfill 
    \begin{subfigure}[b]{\mysubfloatwidththree}
        \fbox{\includegraphics[width=\textwidth]{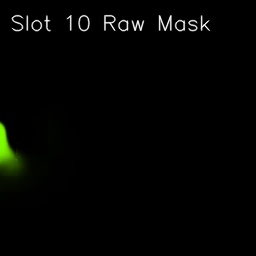}}
    \end{subfigure}
    \hfill 
    \begin{subfigure}[b]{\mysubfloatwidththree}
        \fbox{\includegraphics[width=\textwidth]{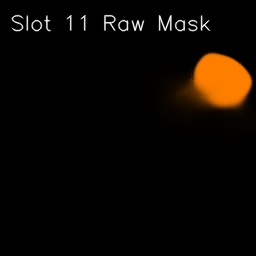}}
    \end{subfigure}
    \hfill 
    \begin{subfigure}[b]{\mysubfloatwidththree}
        \fbox{\includegraphics[width=\textwidth]{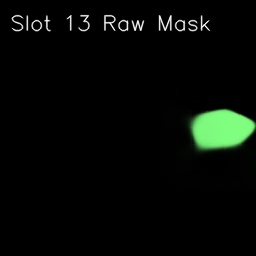}}
    \end{subfigure}
    \hfill 
    \begin{subfigure}[b]{\mysubfloatwidththree}
        \fbox{\includegraphics[width=\textwidth]{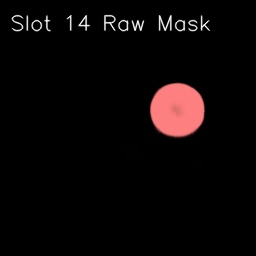}}
    \end{subfigure}
    \hfill 
    \begin{subfigure}[b]{\mysubfloatwidththree}
        \fbox{\includegraphics[width=\textwidth]{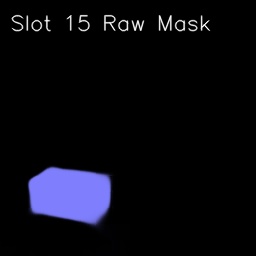}}
    \end{subfigure}
    \hfill 
    \begin{subfigure}[b]{\mysubfloatwidththree}
        \fbox{\includegraphics[width=\textwidth]{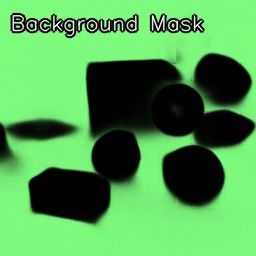}}
    \end{subfigure}
    \\
    
    \begin{subfigure}[b]{\mysubfloatwidththree}
        \fbox{\includegraphics[width=\textwidth]{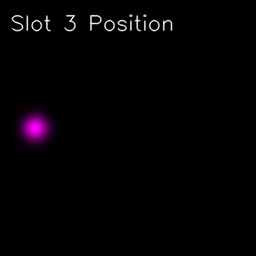}}
    \end{subfigure}
    \hfill 
    \begin{subfigure}[b]{\mysubfloatwidththree}
        \fbox{\includegraphics[width=\textwidth]{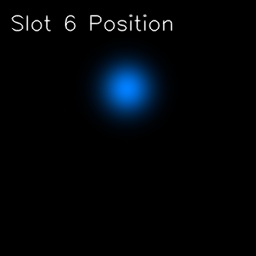}}
    \end{subfigure}
    \hfill 
    \begin{subfigure}[b]{\mysubfloatwidththree}
        \fbox{\includegraphics[width=\textwidth]{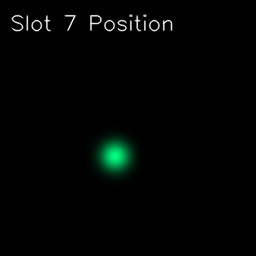}}
    \end{subfigure}
    \hfill 
    \begin{subfigure}[b]{\mysubfloatwidththree}
        \fbox{\includegraphics[width=\textwidth]{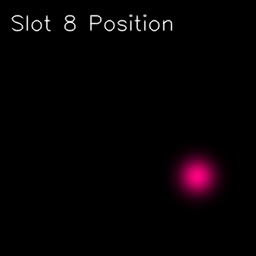}}
    \end{subfigure}
    \hfill 
    \begin{subfigure}[b]{\mysubfloatwidththree}
        \fbox{\includegraphics[width=\textwidth]{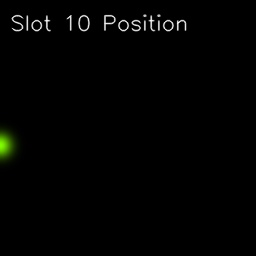}}
    \end{subfigure}
    \hfill 
    \begin{subfigure}[b]{\mysubfloatwidththree}
        \fbox{\includegraphics[width=\textwidth]{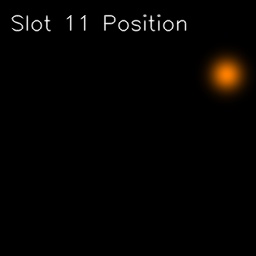}}
    \end{subfigure}
    \hfill 
    \begin{subfigure}[b]{\mysubfloatwidththree}
        \fbox{\includegraphics[width=\textwidth]{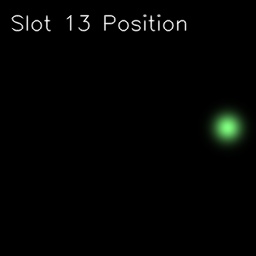}}
    \end{subfigure}
    \hfill 
    \begin{subfigure}[b]{\mysubfloatwidththree}
        \fbox{\includegraphics[width=\textwidth]{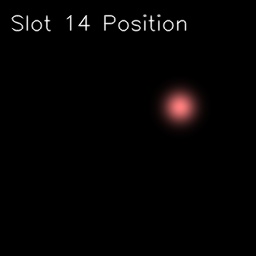}}
    \end{subfigure}
    \hfill 
    \begin{subfigure}[b]{\mysubfloatwidththree}
        \fbox{\includegraphics[width=\textwidth]{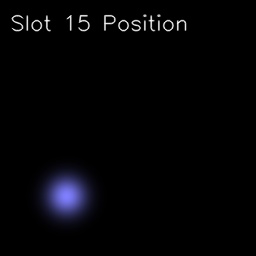}}
    \end{subfigure}
    \hfill 
    \begin{subfigure}[b]{\mysubfloatwidththree}
        \fbox{\includegraphics[width=\textwidth]{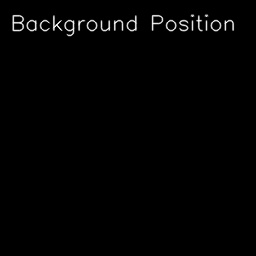}}
    \end{subfigure}
    
    \caption{Example of an internal scene segmentation inferred and processed by \locis*. First row: ground truth RGB image and depth map as well as details of the encoding of the partially occluded large box. Columns in subsequent rows: 
    Interpretable slot-wise decomposition (columns) of the inputs into RGB, depth, and mask reconstructions as well as their position estimates (rows). Only occupied slots (columns) and the background module output (most right column) are shown.
    }
\end{figure*}
In the process of pre-training the encoder-decoder architecture on discrete object instances, the \locis* network acquires a foundational ability to en- and decode objects. Learning is furthermore supported by the pre-trained background module, which distinguishes foreground entities from the background context. The remaining challenge lies in the accurate identification and allocation of objects into distinct slots. 
SAVi++ provides ground-truth bounding box information to accomplish this step \citep{Elsayed:2022}.
Our work explores three methodologies for the initial assignment of slots, without relying on ground-truth information.

First, we utilize a stochastic positioning strategy within the foreground mask that is generated by the Uncertainty Network. 
In particular, we sample a location uniformly randomly within the generated uncertainty map, which essentially predicts object masks. 
We initialize an empty slot with this location encoded as a 2D Gaussian position $Q_k^t$, similar to the pretraining of the object-specific encodings and decodings specified above.

This first approach, however, is susceptible to the erroneous partitioning of larger objects, because multiple random positions may be selected within the same object. 
To mitigate this, our second approach---termed ``Regularized Initial Slots''---retains the random sampling paradigm during a warm-up phase. Following each network pass, though, we compute a similarity metric for each slot pair, based on both the Euclidean distance between their positional codes and the correlation of their Gestalt codes. Slots exhibiting a similarity below a predefined threshold are nullified in a stochastic manner.

The third approach employs a specialized segmentation network akin to YOLACT \citep{bolya2019yolact}, which was trained supervised using a Cross-Entropy loss comparing predicted instance masks with the best matching ground truth once. 
For more details 
see appendix (cf. \autoref{tab:segmentation_unet_architecture} and \autoref{tab:seg_perform}). Initial slot positions are calculated based on the instance masks outputted by this network. 

\begin{table*}[t]
    \centering   
    \caption{\locis* demonstrates largely superior performance in the MOVi Challenge,
    benchmarked against SAVi++ \citep{Elsayed:2022}. 
    The results show that segmentation performance critically depends on the strategy for initial slot assignment. 
    Note that SAVI++ provides ground-truth masks to each slot in the first frame. 
    \locis*, on the other hand, uses our novel segmentation preprocessing strategy. 
    A segmentation network-informed slot assignment (seg) with depth information as input (\locisd*) yields the best score. 
    Random slot assignments given the uncertainty map from the Uncertainty Module (rnd) clearly show the importance to start with good slot assignments, yielding performance worse than SAVi++ but still better than SAVi in MOVi-D and MOVi-E. Employing the regularized initial slot technique (reg) during the initialization phase yields intermediate outcomes. The smaller standard deviation ($\pm$) of our results also hints at a more reproducible training and evaluation of \locis* than SAVI++ and especially SAVI.
    }
    \label{tab:savi_comparison}
    \begin{tabular}{c|ccc|ccc}
        \hline
        & \multicolumn{3}{c|}{mIoU$\uparrow$ (\%)} & \multicolumn{3}{c}{FG-ARI$\uparrow$ (\%)} \\
        Model & MOVi-C & MOVi-D & MOVi-E & MOVi-C & MOVi-D & MOVi-E \\
        \hline
        CRW & \(27.8 \pm 0.2\) & \(45.3 \pm 0.0\) & \(47.5 \pm 0.1\) & * & * & * \\
        SAVi & \(43.1 \pm 0.7\) & \(22.7 \pm 7.5\) & \(30.7 \pm 4.9\) & \(77.6 \pm 0.7\) & \(59.6 \pm 6.7\) & \(55.3 \pm 5.8\) \\
        SAVi++ & \(45.2 \pm 0.1\) & \(48.3 \pm 0.5\) & \(47.1 \pm 1.3\) & \(\textbf{81.9} \pm \textbf{0.2}\) & \(\textbf{86.0} \pm \textbf{0.3}\) & \(84.1 \pm 0.9\) \\
        \midrule
        \locisd* (rnd) & \(40.3 \pm 0.1\) & \(40.9 \pm 0.3\) & \(44.4 \pm 0.2\) & \(58.5 \pm 0.6\) & \(49.8 \pm 0.2\) & \(60.8 \pm 0.2\) \\
        \locisd* (reg) & \(45.7 \pm 0.2\) & \(47.9 \pm 0.3\) & \(49.2 \pm 0.3\) & \(74.1 \pm 0.3\) & \(74.0 \pm 0.9\) & \(81.3 \pm 0.8\) \\
        \locisd* (seg) & \(\textbf{45.5} \pm \textbf{0.1}\) & \(\textbf{51.8} \pm \textbf{0.1}\) & \(\textbf{53.5} \pm \textbf{0.1}\) & \(79.2 \pm 0.3\) & \(81.1 \pm 0.2\) & \(\textbf{88.5} \pm \textbf{0.2}\) \\
        \midrule
        \locis* (rnd) & \(33.4 \pm 0.3\) & \(38.8 \pm 0.2\) & \(41.2 \pm 0.2\) & \(60.4 \pm 0.5\) & \(54.3 \pm 0.3\) & \(63.9 \pm 0.6\) \\
        \locis* (reg) & \(35.7 \pm 0.2\) & \(41.2 \pm 0.2\) & \(42.4 \pm 0.1\) & \(68.1 \pm 0.5\) & \(71.4 \pm 0.7\) & \(78.3 \pm 0.1\) \\
        \locis* (seg) & \(36.2 \pm 0.1\) & \(44.9 \pm 0.1\) & \(47.0 \pm 0.1\) & \(72.7 \pm 0.3\) & \(79.5 \pm 0.2\) & \(85.1 \pm 0.0\) \\

        \hline
    \end{tabular}
\end{table*}

\section{Experiments \& Results}

In our experimentation pipeline, we initially pretrain our models on the Kubric MOVi-(a-f) dataset \citep{greff2022}. Subsequently, we employ two distinct strategies: (1) full model training on MOVi-(a-e) for benchmarking against SAVi++ \citep{Elsayed:2022}, and (2) fine-tuning the encoder, decoder, and background modules on the datasets delineated in the Computational Scene Representation Review \citep{Yuan:2023} prior to training the full \locis* system on these datasets.

For video dataset training, we employ a warm-up phase comprising three iterations, during which only the encoder and decoder are updated, omitting the predictor. This warm-up occurs on the initial frame. We utilize truncated backpropagation through time (BPTT) with a sequence length of 2 for sequence-based learning. In contrast, for image datasets, we omit the warm-up phase and iteratively forward and backward propagate the same image for three cycles.

In the inference phase, we extend the warm-up iterations to 10 for both video and image data, which showed an empirical improvement in performance. Additionally, in image-centric tasks, we augment the number of full-architecture iterations to 10, totaling 20 processing steps: 10 for encoder/decoder-only warm-up and 10 for full architecture evaluation. 

\subsection{Video Evaluation}

The design of \locis* primarily centers around temporal predictions, hence a detailed comparative study is performed against SAVi++ on the MOVi-(c-e) dataset. To maintain evaluative consistency, we adhere to the same performance measures as outlined in the SAVi++ paper, namely the Per-Sequence Intersection over Union (IoU) and the Per-Sequence Foreground Adjusted Rand Index (FG-ARI). 

As illustrated in \autoref{tab:savi_comparison}, \locis* manifests a notable performance uplift, registering a 13.59\% relative IoU
improvement on the most demanding MOVi-E dataset, elevating the score from 47.1\% (attained by SAVi++) to 53.5\% (\locis* with depth input and segmentation preprocessing). The FG-ARI results are more mixed with \locis* achieving superior performance in the most challenging MOVi-E dataset while achieving slightly lower scores than SAVi++ on MOVi-C and MOVi-D.

However, an important consideration in interpreting the discrepancy between IoU and FG-ARI scores involves recognizing the inherent nature of these metrics, particularly in the scope of temporal segmentation. IoU, in this context, can be interpreted as a 'worst-case' metric. It fundamentally penalizes any misalignment between predicted and ground truth instances, inherently emphasizing the lower bound of segmentation accuracy. This stringent criterion means that even minor deviations in spatial alignment or instance misidentification across the sequence are heavily penalized, making high IoU scores indicative of exceptional spatial and temporal precision in segmentation.

Conversely, the FG-ARI, by design, is a metric that emphasizes the consistency and correctness of instance tracking over time, offering a more nuanced perspective on temporal segmentation. It quantifies the accuracy of instance groupings while allowing some leeway for minor errors in boundary delineation or slight temporal shifts. This characteristic of FG-ARI implies that, despite a lower score in this metric, \locis* may still effectively capture the overall dynamics and structure of the scene over time while performing slightly worse in object tracking than SAVi++ on MOVi-C and MOVi-D.
\begin{table*}[t]
\centering
\caption{Comparative results of \locis* across six datasets for compositional scene understanding (moving MNIST digits and dSprites, the Abstract Scene dataset, CLEVR, SHOP VRB, and a combination of GSO
and HDRI-Haven) \citep{Yuan:2023}. The table shows in-distribution performance (3-6 objects) and out-of-distribution generalization (7-10 objects). Models were pretrained on MOVi-(a-f), fine-tuned on all datasets, and evaluated using the epoch with lowest validation error. Generalization capacity was assessed by increasing the maximum number of slots to 10 without further training. \locis* largely outperforms all other reported approaches.}
\label{tab:review}
\begin{tabular}{l|cccccccc}
\hline
 & AMI-A & ARI-A & AMI-O & ARI-O & IoU & F1 & OCA & OOA \\
\hline
\multicolumn{9}{c}{Test 1 Validation (3-6 Objects)} \\
\hline
  AIR & 0.380 & 0.397 & 0.845 & 0.827 & N/A & N/A & 0.549 & 0.709 \\
  N-EM & 0.208 & 0.233 & 0.341 & 0.282 & N/A & N/A & 0.013 & N/A \\
  IODINE & 0.638 & 0.700 & 0.772 & 0.752 & N/A & N/A & 0.487 & N/A \\
  GMIOO & 0.738 & 0.811 & \textbf{0.916} & 0.914 & 0.708 & 0.808 & \textbf{0.772} & \textbf{0.846} \\
  MONet & 0.657 & 0.699 & 0.863 & 0.857 & N/A & N/A & 0.663 & 0.583 \\
  GENESIS & 0.411 & 0.412 & 0.420 & 0.382 & 0.105 & 0.170 & 0.213 & 0.603 \\
  SPACE & 0.640 & 0.678 & 0.817 & 0.765 & 0.630 & 0.739 & 0.436 & 0.666 \\
  Slot Attention & 0.393 & 0.321 & 0.758 & 0.711 & N/A & N/A & 0.028 & N/A \\
  EfficientMORL & 0.341 & 0.279 & 0.673 & 0.621 & N/A & N/A & 0.107 & N/A \\
  GENESIS-V2 & 0.304 & 0.206 & 0.728 & 0.693 & N/A & N/A & 0.153 & 0.574 \\
  \hline
  \locis* (rnd) & 0.835 & 0.918 & 0.735 & 0.891 & 0.742 & 0.822 & 0.421 & - \\
  \locis* (reg) & 0.838 & 0.921 & 0.730 & 0.898 & 0.731 & 0.810 & 0.443 & - \\
  \locis* (seg) & \textbf{0.844} & \textbf{0.922} & 0.748 & \textbf{0.920} & \textbf{0.781} & \textbf{0.860} & 0.505 & - \\
  \hline
\hline
\multicolumn{9}{c}{Test 2 Generalization (7-10 Objects)} \\
\hline
  AIR & 0.410 & 0.402 & 0.802 & 0.740 & N/A & N/A & 0.327 & 0.689 \\
  N-EM & 0.256 & 0.268 & 0.354 & 0.261 & N/A & N/A & 0.017 & N/A \\
  IODINE & 0.633 & 0.652 & 0.781 & 0.731 & N/A & N/A & 0.387 & N/A \\
  GMIOO & 0.732 & 0.781 & \textbf{0.891} & 0.868 & 0.647 & 0.746 & \textbf{0.534} & \textbf{0.823} \\
  MONet & 0.635 & 0.665 & 0.820 & 0.785 & N/A & N/A & 0.446 & 0.619 \\
  GENESIS & 0.380 & 0.378 & 0.415 & 0.315 & 0.076 & 0.132 & 0.160 & 0.584 \\
  SPACE & 0.628 & 0.639 & 0.802 & 0.717 & 0.543 & 0.654 & 0.265 & 0.650 \\
  Slot Attention & 0.447 & 0.330 & 0.761 & 0.696 & N/A & N/A & 0.029 & N/A \\
  EfficientMORL & 0.366 & 0.236 & 0.662 & 0.562 & N/A & N/A & 0.085 & N/A \\
  GENESIS-V2 & 0.378 & 0.235 & 0.723 & 0.655 & N/A & N/A & 0.189 & 0.617 \\
  \hline
  \locis* (rnd)       & 0.820 & 0.866 & 0.766 & 0.875 & 0.667 & 0.755 & 0.228 & - \\
  \locis* (reg)  & 0.828 & \textbf{0.888} & 0.768 & 0.865 & 0.637 & 0.724 & 0.244 & - \\
  \locis* (seg) & \textbf{0.832} & 0.877 & 0.783 & \textbf{0.905} & \textbf{0.706} & \textbf{0.792} & 0.315 & - \\
\hline
\end{tabular}
\end{table*}

In light of these architectural design choices, \loci* has demonstrated stable slot activations over extended temporal windows \citep{Traub:2023}. 
However, the masks decoded from the predicted Gestalt codes ${M}_k^{t+1}$ are less accurate than those derived directly from the encoder $\tilde M_k^{t}$. The encoder masks, on the other hand, suffer from low temporal consistency, since the information fusion of $\tilde{G}_k^t$ with $G_k^{t-1}$ happens afterwards via the update gate and also inside the predictor itself (via the GateL0RD recurrences). At the moment the challenge remains to further improve the fusion of accurate mask reconstructions ($\tilde G_k^t$) with stable temporal predictions ($\tilde G_k^{t-1}$).

\subsection{Image Evaluation}
\definecolor{slotpanelcolor}{RGB}{140, 160, 190}
\colorlet{boxplotfillcolor}{slotpanelcolor!40!white}

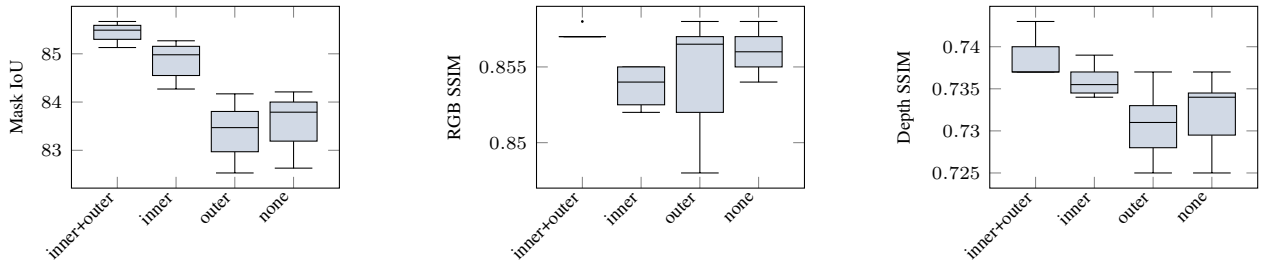
\begin{figure*}[t]
  \centering
  \begin{subfigure}[b]{0.3\textwidth}
    \centering
      \begin{tikzpicture}
    \begin{axis}[
		boxplot/draw direction=y,
		ylabel={Mask IoU},
        xlabel={},
		height=4cm,
        xtick={1,2,3,4},
		xticklabels={inner+outer,inner,outer,none},
        xticklabel style={rotate=45, anchor=east},
		style={font=\scriptsize},
      	width=\textwidth,
    ]

\addplot+[blackboxplot,fill=boxplotfillcolor] table [row sep=\\, y index=0] {
              data\\
85.670000000000002\\
85.670000000000002\\
85.129999999999995\\
85.510000000000005\\
85.469999999999999\\};
\addplot+[blackboxplot,fill=boxplotfillcolor] table [row sep=\\, y index=0] {
              data\\
84.829999999999998\\
85.269999999999996\\
85.180000000000007\\
84.269999999999996\\
85.129999999999995\\};
\addplot+[blackboxplot,fill=boxplotfillcolor] table [row sep=\\, y index=0] {
              data\\
82.530000000000001\\
83.409999999999997\\
83.530000000000001\\
84.170000000000002\\
84.079999999999998\\};
\addplot+[blackboxplot,fill=boxplotfillcolor] table [row sep=\\, y index=0] {
              data\\
83.750000000000000\\
82.629999999999995\\
84.209999999999994\\
83.829999999999998\\
84.170000000000002\\};

    \end{axis}
  \end{tikzpicture}

  \end{subfigure}
  \hfill
  \begin{subfigure}[b]{0.3\textwidth}
    \centering
  \begin{tikzpicture}
    \begin{axis}[
		boxplot/draw direction=y,
		ylabel={RGB SSIM},
        xlabel={},
		height=4cm,
        yticklabel style={
          /pgf/number format/fixed,
          /pgf/number format/precision=3
        },
        xtick={1,2,3,4},
		xticklabels={inner+outer,inner,outer,none},
        xticklabel style={rotate=45, anchor=east},
		style={font=\scriptsize},
      	width=\textwidth,
    ]

\addplot+[blackboxplot,fill=boxplotfillcolor] table [row sep=\\, y index=0] {
data\\
0.857\\
0.858\\
0.857\\
0.857\\
0.857\\
};
\addplot+[blackboxplot,fill=boxplotfillcolor] table [row sep=\\, y index=0] {
data\\
0.852\\
0.855\\
0.855\\
0.853\\
0.855\\
};
\addplot+[blackboxplot,fill=boxplotfillcolor] table [row sep=\\, y index=0] {
data\\
0.848\\
0.856\\
0.857\\
0.857\\
0.858\\
};
\addplot+[blackboxplot,fill=boxplotfillcolor] table [row sep=\\, y index=0] {
data\\
0.856\\
0.854\\
0.858\\
0.856\\
0.858\\
};

    \end{axis}
  \end{tikzpicture}
    
  \end{subfigure}
  \hfill
  \begin{subfigure}[b]{0.3\textwidth}
    \centering
  \begin{tikzpicture}
    \begin{axis}[
		boxplot/draw direction=y,
		ylabel={Depth SSIM},
        xlabel={},
		height=4cm,
        yticklabel style={
          /pgf/number format/fixed,
          /pgf/number format/precision=3
        },
        xtick={1,2,3,4},
		xticklabels={inner+outer,inner,outer,none},
        xticklabel style={rotate=45, anchor=east},
		style={font=\scriptsize},
      	width=\textwidth,
    ]

\addplot+[blackboxplot,fill=boxplotfillcolor] table [row sep=\\, y index=0] {
data\\
0.743\\
0.743\\
0.737\\
0.737\\
0.737\\
};
\addplot+[blackboxplot,fill=boxplotfillcolor] table [row sep=\\, y index=0] {
data\\
0.736\\
0.738\\
0.739\\
0.734\\
0.735\\
};
\addplot+[blackboxplot,fill=boxplotfillcolor] table [row sep=\\, y index=0] {
data\\
0.725\\
0.731\\
0.731\\
0.737\\
0.735\\
};
\addplot+[blackboxplot,fill=boxplotfillcolor] table [row sep=\\, y index=0] {
data\\
0.734\\
0.725\\
0.737\\
0.734\\
0.735\\
};
    \end{axis}
  \end{tikzpicture}
    
  \end{subfigure}
  \caption{Ablating inner or outer feedback loop confirms their efficacy seeing 
  improvements in per-object mask intersection-over-union (IoU) and depth structural similarity index measure (SSIM).
  }
  \label{fig:hyper_ablation}
\end{figure*}
In a recent review paper about compositional scene understanding, \citet{Yuan:2023} proposed a total of 6 datasets ranging in complexity from compositing MNIST to realistic texture simulations like MOVi-(c-e). These datasets are constructed in a way to perform two test: an in-distribution test with the same number of objects in a scene as seen during training (between 3 and 6); and another out-of-distribution test that probes generalization abilities with object numbers ranging from 7 to 10. In our experiments we used a pretraind (on MOVi-(a-f)) encoder-decoder network and fine-tuned it using all 6 datasets at once. We then further trained \locis* on these combined 6 datasets and set the maximum number of slots to 6 during training. We then selected the model checkpoint form the epoch with the lowest validation error. 

For the generalization test we simply increase the maximum number of slots without further training to 10. We test the following metrics, which were reported by \cite{Yuan:2023}: Adjusted Mutual Information (AMI), Adjusted Rand Index (ARI), Intersection over Union (IoU), F1 score and Object Counting Accuracy (OCA). As shown in \autoref{tab:review} \locis* shows superior performance in most metrics for both the in-distribution test and the generalization test. Note that we did not explicitly train or fine-tune \locis* on each dataset individually, as don in \citet{Yuan:2023}, but rather trained them on all datasets combined, scaling individual resolutions up to $256\times 256$ where necessary. We expect even better performance with a dataset-specific fine-tuning of model and hyper parameters.

\subsection{Top Down Feedback Ablations}

In \autoref{fig:hyper_ablation}, we conduct a further ablation study to investigate the impact of top-down feedback in our architecture. We restrict our evaluation to pre-trained encoder-decoder networks, motivated by their substantially lower computational cost. Indeed, these networks are trainable using a single GTX 1080 GPU with a single slot for pre-training. Each experimental configuration is executed five times, utilizing a consistent set of five random seeds for reproducibility. Our ablation compares four scenarios evaluating the inner and outer top-down feedback loops: the inner feedback controls the hyper-network tuning the spatial convolutional encoder kernels top-down using $\hat G^t_k$;
the outer feedback refers to the slot specific inputs $\hat R^t_k$, $\hat D^t_k$, $M^{t,o}_k$, $\hat M^{t,v}_k$ and $\hat M^{t,s}_k$.
We compare the performance of 
(i) the proposed architecture with both inner and outer feedback loops, (ii) a version with only outer feedback, (iii) a version with only inner feedback, and (iv) a baseline with no feedback mechanisms.

The results in \autoref{fig:hyper_ablation} demonstrate that the inclusion of top-down feedback is particularly advantageous for mask prediction tasks, resulting in a significant improvement in the Intersection-over-Union (IoU) metric. While the benefits for the Structural Similarity Index Measure (SSIM) in RGB reconstruction are less consistent, the depth reconstruction task also profits from the presence of top-down feedback information.

\section{Conclusion}

The \locis* model introduces several key innovations in the domain of scene understanding and object segmentation. A novel background reconstruction and foreground density estimation approach greatly facilitates object-oriented scene segmentations without relying on ground-truth slot initialization. 
Moreover, dynamic convolution kernels via a hyper-network-controlled top-down residual network facilitates object-specific visual encoding. 
Finally, the incorporation of depth information additionally facilitates segmentation performance event further. 
These advancements collectively contribute to a 13.59\% relative improvement in IoU on the challenging MOVi-E dataset compared to state-of-the-art models like SAVi++. 
Still, \locis* falls short in some performance metrics, particularly in FG-ARI in the MOVi-C and MOVi-D datasets when compared to SAVi++. 
This suggests that while \locis* excels in segmenting complex environments, it still struggles with fully accurate temporal object trackings. 
We suspect that this can be attributed to the fact that \locis* fully compresses past video frame information in the internal recurrent state of its Transition Module. 
Our results demonstrate robustness in both in-distribution and out-of-distribution tests, highlighting the model's generalization abilities. However, it remains an open question whether this robustness extends to more varied or even more dynamic environments, and how it fares against models optimized for such scenarios.
Furthermore, \locis* shows great potential in terms of interpretability of deep learning systems, as shown in \autoref{fig:occlusion} (further examples can be found in the appendix and supplementary video material).
 
Taking inspiration from human cognition, from the binding problem, and from recent computational and conceptual insights into our modularized minds \citep{Greff:2020,Mattar:2022,Heald:2023,Butz:2021epc,Schwoebel:2021}, the background processing module may yet be enhanced to a universal background extraction module relative to which foreground objects may be extracted. 
Furthermore, optimally distributing cognitive processing resources onto currently task-relevant objects and interactions remains as an important challenge. 
We believe that segmentation-oriented algorithms, such as \locis*, constitute one crucial foundation-model-like module that offers itself to be effectively combined with (i) reinforcement learning, planning, and reasoning approaches and (ii) language processing modules in future work.


\section{Acknowledgement}
This work received funding from the Deutsche Forschungsgemeinschaft (DFG, German Research Foundation) under Germany’s Excellence Strategy – EXC number 2064/1 – Project number 390727645 as well as from the Cyber Valley in Tübingen, CyVy-RF-2020-15. The authors thank the International Max Planck Research School for Intelligent Systems (IMPRS-IS) for supporting Manuel Traub, and the Alexander von Humboldt Foundation for supporting Martin Butz and Sebastian Otte.

\bibliography{iclr2024_conference}
\bibliographystyle{icml2024}

\appendix
\section{Appendix}

\subsection{Closing the Inner Loop}
We enhance \loci*'s object tracking abilities similar to \cite{traub2023looping}, which draws inspiration from Kalman filtering \citep{kalman_new_1960}. 
Originally, \loci* predicts the next object states via a pixel space-routed outer loop (see \autoref{fig:loci}; outer loop). We draw inspiration from work in model-based reinforcement learning, which has recently advocated latent world model predictions \citep{hafner_learning_2019,hafner_dream_2020, ha_world_2018,Schrittwieser:2020}. These allow the imagination of future scene dynamics via an inner loop, without explicit pixel-based generations. 
Similarly, we apply an inner processing loop in \locis*.
(see \autoref{fig:loci}; inner loop). 

In accordance with Kalman filtering, \locis* is enabled to linearly interpolate between the current sensor information and its predictions. Formally, the current object states $S^{t}_k=(\smash{G^{t}_k}$, $\smash{P^{t}_k})$ become a linear blending of the observed object states $\smash{\tilde{G}^{t}_k}$, $\smash{\tilde{P}^{t}_k}$ and the predicted object states $\smash{\hat{G}^{t}_k}$, $\smash{\hat{P}^{t}_k}$:
\begin{gather}
  G^{t}_k = \alpha^{t,G}_k \tilde{G}^{t}_k + (1-\alpha^{t,G}_k) \hat{G}^{t}_k  
\\
 P^{t}_k = \alpha^{t,P}_k \tilde{P}^{t}_k + (1-\alpha^{t,P}_k) \hat{P}^{t}_k
\end{gather}
The weighting $\alpha$ is specific for each Gestalt and position code in each slot $k$. Importantly, \locis* learns to regulate this percept gate on its own in a fully self-supervised manner. It learns an update function $g_{\theta}$, which takes as input the observed state $\tilde{S}^{t}_k$, the predicted state $\hat{S}^{t}_k$, and the last positional encoding $P^{t-1}_k$:
\begin{equation}
    (z_k^{t,G}, z_k^{t,P}) = g_{\theta}(\tilde{S}^{t}_k, \hat{S}^{t}_k, P^{t-1}_k) + \varepsilon \hspace{0.5in} \text{with} \hspace{0.1in} \varepsilon \sim \mathcal{N}(0, \Sigma) ,
\end{equation}
We model $g_{\theta}$ with a feed-forward network. To be able to fully rely on its own predictions, \locis* needs to be able to fully close the gate by setting $\alpha$ exactly to zero. We therefore use a rectified hyperbolic tangent to compute $\alpha$:
\begin{equation}
    (\alpha^{t,G}_k, \alpha^{t,P}_k) = \mathrm{max}(0, \mathrm{tanh}((z_k^{t,G}, z_k^{t,P}))).
\end{equation}
An $L_0$ loss on gate openings encourages the reliance on internal beliefs rather than external updates.

\section{Depth Input Normalization}
\label{app:depth_norm}
We log-normalized the Scene-Relative Depth, according to \autoref{eq:depth_norm}:
\begin{equation}\label{eq:depth_norm}
d = \frac{1}{1 + \exp\left(\frac{\hat{d} - \mu}{\sigma}\right)},
\end{equation}
where $\hat d$ represents the natural logarithm of the raw depth values. Parameters $\mu$ and $\sigma$ denote the mean and standard deviation of the log-transformed depth, respectively.

\section{Additional tables and figures}

\begin{table*}[t]
    \centering  
    
    \caption{Performance of our segmentation prepossessing network. While achieving adequate performance on evaluation datasets, the preprocessing network clearly fails on the generalization dataset.}
    \label{tab:seg_perform}
    \begin{tabular}{|ccccc|}
        \hline
        \multicolumn{5}{|c|}{mIoU$\uparrow$ (\%)} \\
        MOVi-C & MOVi-D & MOVi-E & Review datasets Test 1 & Review datasets Test 2  \\
        \hline
        \(88.39 \pm 0.03\) & \(82.48 \pm 0.05\) & \(80.89 \pm 0.07\) & \(90.98 \pm 0.12\) & \(67.31 \pm 0.05 \) \\
        \hline
    \end{tabular}
\end{table*}
    
\begin{figure*}[ht]
    \centering
    \setlength{\fboxsep}{0pt}
    \setlength{\fboxrule}{1pt}
    
    \newlength{\mysubfloatwidth}
    \setlength{\mysubfloatwidth}{\dimexpr(\linewidth - 6\fboxrule - 6\fboxsep)/11\relax}

    \begin{subfigure}[b]{\mysubfloatwidth}
        \fbox{\includegraphics[width=\textwidth]{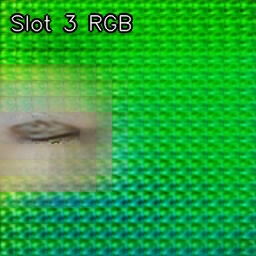}}
    \end{subfigure}
    \hfill 
    \begin{subfigure}[b]{\mysubfloatwidth}
        \fbox{\includegraphics[width=\textwidth]{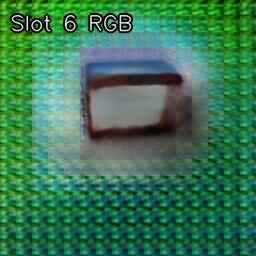}}
    \end{subfigure}
    \hfill 
    \begin{subfigure}[b]{\mysubfloatwidth}
        \fbox{\includegraphics[width=\textwidth]{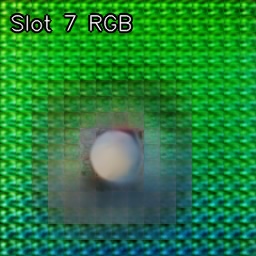}}
    \end{subfigure}
    \hfill 
    \begin{subfigure}[b]{\mysubfloatwidth}
        \fbox{\includegraphics[width=\textwidth]{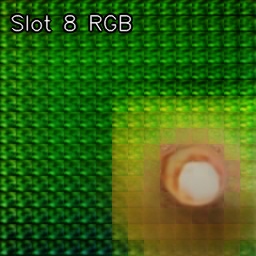}}
    \end{subfigure}
    \hfill 
    \begin{subfigure}[b]{\mysubfloatwidth}
        \fbox{\includegraphics[width=\textwidth]{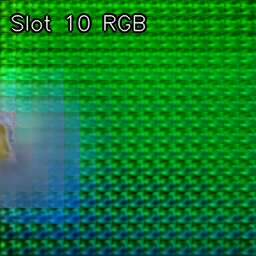}}
    \end{subfigure}
    \hfill 
    \begin{subfigure}[b]{\mysubfloatwidth}
        \fbox{\includegraphics[width=\textwidth]{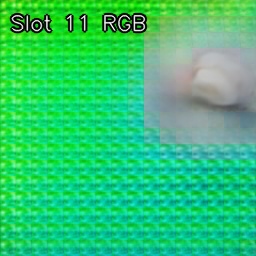}}
    \end{subfigure}
    \hfill 
    \begin{subfigure}[b]{\mysubfloatwidth}
        \fbox{\includegraphics[width=\textwidth]{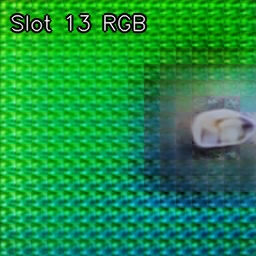}}
    \end{subfigure}
    \hfill 
    \begin{subfigure}[b]{\mysubfloatwidth}
        \fbox{\includegraphics[width=\textwidth]{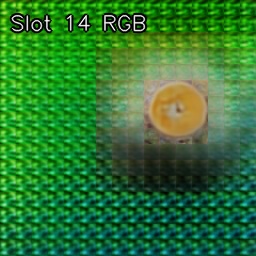}}
    \end{subfigure}
    \hfill 
    \begin{subfigure}[b]{\mysubfloatwidth}
        \fbox{\includegraphics[width=\textwidth]{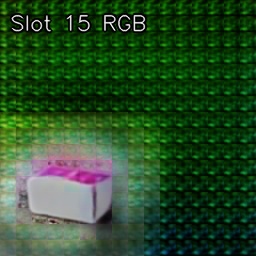}}
    \end{subfigure}
    \hfill 
    \begin{subfigure}[b]{\mysubfloatwidth}
        \fbox{\includegraphics[width=\textwidth]{imgs/object-0005-033-obj16.jpg}}
    \end{subfigure}
    \\

    \begin{subfigure}[b]{\mysubfloatwidth}
        \fbox{\includegraphics[width=\textwidth]{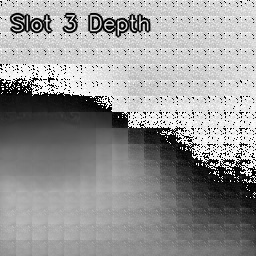}}
    \end{subfigure}
    \hfill 
    \begin{subfigure}[b]{\mysubfloatwidth}
        \fbox{\includegraphics[width=\textwidth]{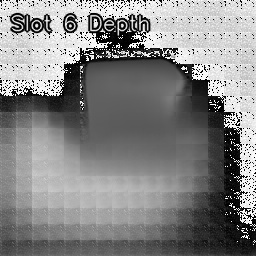}}
    \end{subfigure}
    \hfill 
    \begin{subfigure}[b]{\mysubfloatwidth}
        \fbox{\includegraphics[width=\textwidth]{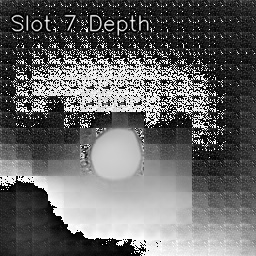}}
    \end{subfigure}
    \hfill 
    \begin{subfigure}[b]{\mysubfloatwidth}
        \fbox{\includegraphics[width=\textwidth]{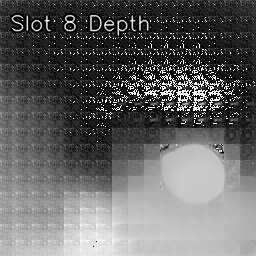}}
    \end{subfigure}
    \hfill 
    \begin{subfigure}[b]{\mysubfloatwidth}
        \fbox{\includegraphics[width=\textwidth]{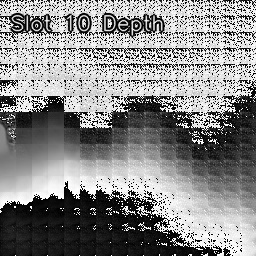}}
    \end{subfigure}
    \hfill 
    \begin{subfigure}[b]{\mysubfloatwidth}
        \fbox{\includegraphics[width=\textwidth]{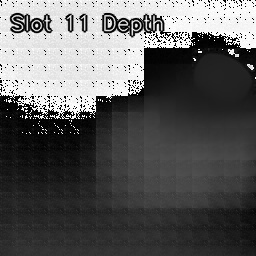}}
    \end{subfigure}
    \hfill 
    \begin{subfigure}[b]{\mysubfloatwidth}
        \fbox{\includegraphics[width=\textwidth]{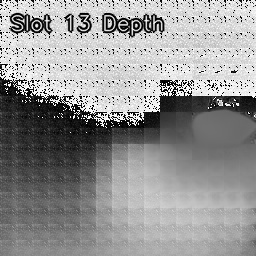}}
    \end{subfigure}
    \hfill 
    \begin{subfigure}[b]{\mysubfloatwidth}
        \fbox{\includegraphics[width=\textwidth]{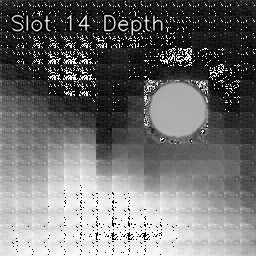}}
    \end{subfigure}
    \hfill 
    \begin{subfigure}[b]{\mysubfloatwidth}
        \fbox{\includegraphics[width=\textwidth]{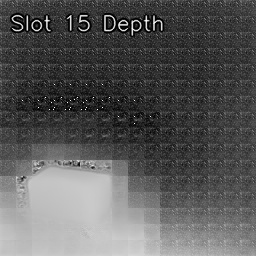}}
    \end{subfigure}
    \hfill 
    \begin{subfigure}[b]{\mysubfloatwidth}
        \fbox{\includegraphics[width=\textwidth]{imgs/depth-0005-033-obj16.jpg}}
    \end{subfigure}
    \\
    
    \begin{subfigure}[b]{\mysubfloatwidth}
        \fbox{\includegraphics[width=\textwidth]{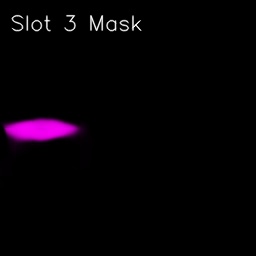}}
    \end{subfigure}
    \hfill 
    \begin{subfigure}[b]{\mysubfloatwidth}
        \fbox{\includegraphics[width=\textwidth]{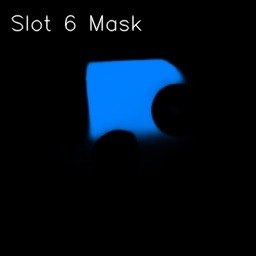}}
    \end{subfigure}
    \hfill 
    \begin{subfigure}[b]{\mysubfloatwidth}
        \fbox{\includegraphics[width=\textwidth]{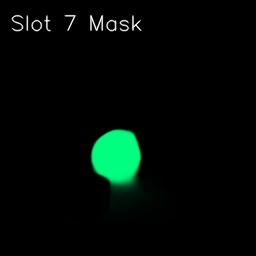}}
    \end{subfigure}
    \hfill 
    \begin{subfigure}[b]{\mysubfloatwidth}
        \fbox{\includegraphics[width=\textwidth]{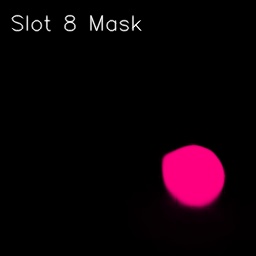}}
    \end{subfigure}
    \hfill 
    \begin{subfigure}[b]{\mysubfloatwidth}
        \fbox{\includegraphics[width=\textwidth]{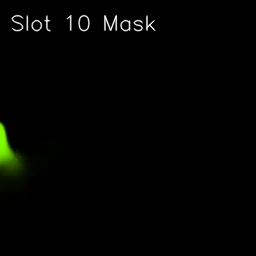}}
    \end{subfigure}
    \hfill 
    \begin{subfigure}[b]{\mysubfloatwidth}
        \fbox{\includegraphics[width=\textwidth]{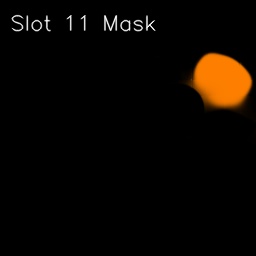}}
    \end{subfigure}
    \hfill 
    \begin{subfigure}[b]{\mysubfloatwidth}
        \fbox{\includegraphics[width=\textwidth]{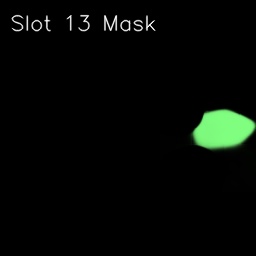}}
    \end{subfigure}
    \hfill 
    \begin{subfigure}[b]{\mysubfloatwidth}
        \fbox{\includegraphics[width=\textwidth]{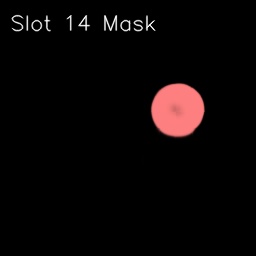}}
    \end{subfigure}
    \hfill 
    \begin{subfigure}[b]{\mysubfloatwidth}
        \fbox{\includegraphics[width=\textwidth]{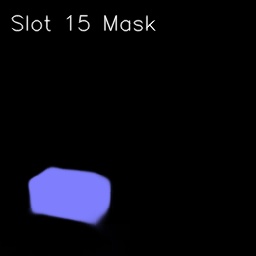}}
    \end{subfigure}
    \hfill 
    \begin{subfigure}[b]{\mysubfloatwidth}
        \fbox{\includegraphics[width=\textwidth]{imgs/mask-0005-033-obj16.jpg}}
    \end{subfigure}
    \\
    
    \begin{subfigure}[b]{\mysubfloatwidth}
        \fbox{\includegraphics[width=\textwidth]{imgs/position-0005-033-obj03.jpg}}
    \end{subfigure}
    \hfill 
    \begin{subfigure}[b]{\mysubfloatwidth}
        \fbox{\includegraphics[width=\textwidth]{imgs/position-0005-033-obj06.jpg}}
    \end{subfigure}
    \hfill 
    \begin{subfigure}[b]{\mysubfloatwidth}
        \fbox{\includegraphics[width=\textwidth]{imgs/position-0005-033-obj07.jpg}}
    \end{subfigure}
    \hfill 
    \begin{subfigure}[b]{\mysubfloatwidth}
        \fbox{\includegraphics[width=\textwidth]{imgs/position-0005-033-obj08.jpg}}
    \end{subfigure}
    \hfill 
    \begin{subfigure}[b]{\mysubfloatwidth}
        \fbox{\includegraphics[width=\textwidth]{imgs/position-0005-033-obj10.jpg}}
    \end{subfigure}
    \hfill 
    \begin{subfigure}[b]{\mysubfloatwidth}
        \fbox{\includegraphics[width=\textwidth]{imgs/position-0005-033-obj11.jpg}}
    \end{subfigure}
    \hfill 
    \begin{subfigure}[b]{\mysubfloatwidth}
        \fbox{\includegraphics[width=\textwidth]{imgs/position-0005-033-obj13.jpg}}
    \end{subfigure}
    \hfill 
    \begin{subfigure}[b]{\mysubfloatwidth}
        \fbox{\includegraphics[width=\textwidth]{imgs/position-0005-033-obj14.jpg}}
    \end{subfigure}
    \hfill 
    \begin{subfigure}[b]{\mysubfloatwidth}
        \fbox{\includegraphics[width=\textwidth]{imgs/position-0005-033-obj15.jpg}}
    \end{subfigure}
    \hfill 
    \begin{subfigure}[b]{\mysubfloatwidth}
        \fbox{\includegraphics[width=\textwidth]{imgs/position-0005-033-obj16.jpg}}
    \end{subfigure}
    
    \caption{The slotwise decomposition of the input from \autoref{fig:occlusion} into unmasked rgb and depth reconstructions per slot and the background rgb and depth reconstruction. For simplicity we only show occupied slots.}
\end{figure*}

\begin{figure*}[t]
    \centering
    \setlength{\fboxsep}{0pt}
    \setlength{\fboxrule}{1pt}
    
    \newlength{\mysubfloatwidthtwo}
    \setlength{\mysubfloatwidthtwo}{\dimexpr(\linewidth+30pt)/7\relax}

    \begin{subfigure}[b]{\mysubfloatwidthtwo}
        \fbox{\includegraphics[width=\textwidth]{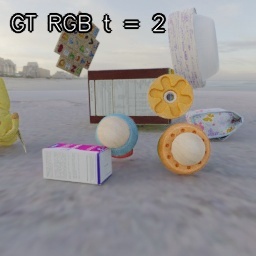}}
    \end{subfigure}
    \hfill 
    \begin{subfigure}[b]{\mysubfloatwidthtwo}
        \fbox{\includegraphics[width=\textwidth]{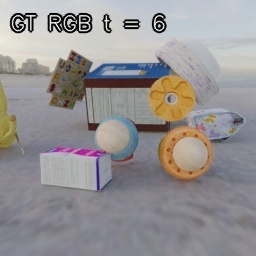}}
    \end{subfigure}
    \hfill 
    \begin{subfigure}[b]{\mysubfloatwidthtwo}
        \fbox{\includegraphics[width=\textwidth]{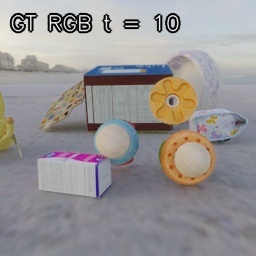}}
    \end{subfigure}
    \hfill 
    \begin{subfigure}[b]{\mysubfloatwidthtwo}
        \fbox{\includegraphics[width=\textwidth]{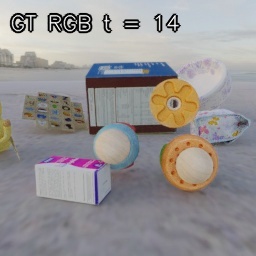}}
    \end{subfigure}
    \hfill 
    \begin{subfigure}[b]{\mysubfloatwidthtwo}
        \fbox{\includegraphics[width=\textwidth]{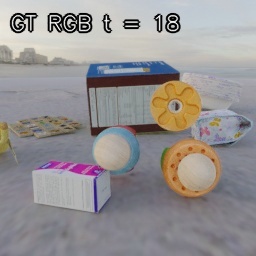}}
    \end{subfigure}
    \hfill 
    \begin{subfigure}[b]{\mysubfloatwidthtwo}
        \fbox{\includegraphics[width=\textwidth]{imgs/input_rgb-0005-033.jpg}}
    \end{subfigure}
    \\
    \vspace{2.45pt}
    \begin{subfigure}[b]{\mysubfloatwidthtwo}
        \fbox{\includegraphics[width=\textwidth]{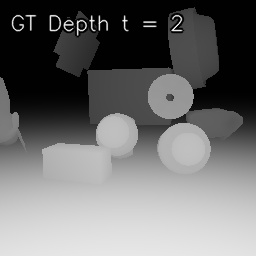}}
    \end{subfigure}
    \hfill 
    \begin{subfigure}[b]{\mysubfloatwidthtwo}
        \fbox{\includegraphics[width=\textwidth]{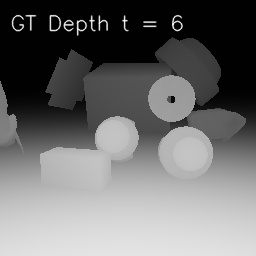}}
    \end{subfigure}
    \hfill 
    \begin{subfigure}[b]{\mysubfloatwidthtwo}
        \fbox{\includegraphics[width=\textwidth]{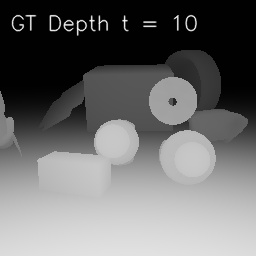}}
    \end{subfigure}
    \hfill 
    \begin{subfigure}[b]{\mysubfloatwidthtwo}
        \fbox{\includegraphics[width=\textwidth]{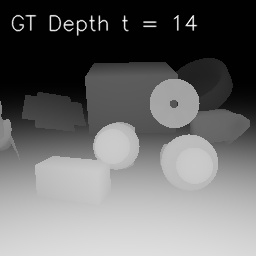}}
    \end{subfigure}
    \hfill 
    \begin{subfigure}[b]{\mysubfloatwidthtwo}
        \fbox{\includegraphics[width=\textwidth]{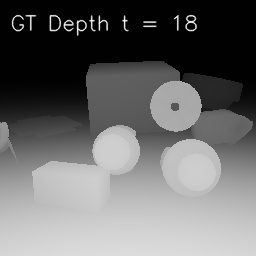}}
    \end{subfigure}
    \hfill 
    \begin{subfigure}[b]{\mysubfloatwidthtwo}
        \fbox{\includegraphics[width=\textwidth]{imgs/input_depth-0005-033.jpg}}
    \end{subfigure}
    \\
    \vspace{2.45pt}
    \begin{subfigure}[b]{\mysubfloatwidthtwo}
        \fbox{\includegraphics[width=\textwidth]{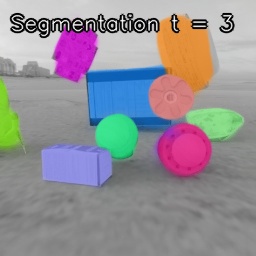}}
    \end{subfigure}
    \hfill 
    \begin{subfigure}[b]{\mysubfloatwidthtwo}
        \fbox{\includegraphics[width=\textwidth]{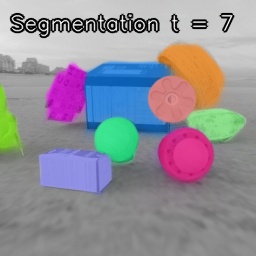}}
    \end{subfigure}
    \hfill 
    \begin{subfigure}[b]{\mysubfloatwidthtwo}
        \fbox{\includegraphics[width=\textwidth]{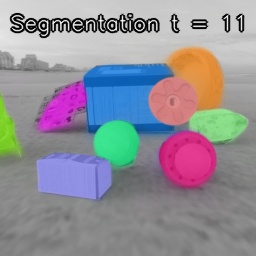}}
    \end{subfigure}
    \hfill 
    \begin{subfigure}[b]{\mysubfloatwidthtwo}
        \fbox{\includegraphics[width=\textwidth]{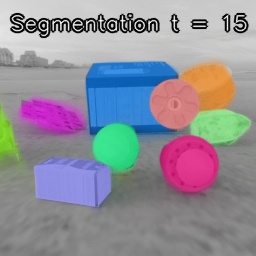}}
    \end{subfigure}
    \hfill 
    \begin{subfigure}[b]{\mysubfloatwidthtwo}
        \fbox{\includegraphics[width=\textwidth]{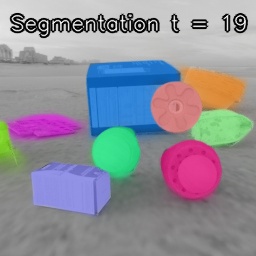}}
    \end{subfigure}
    \hfill 
    \begin{subfigure}[b]{\mysubfloatwidthtwo}
        \fbox{\includegraphics[width=\textwidth]{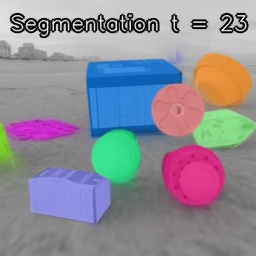}}
    \end{subfigure}
    \\
    \vspace{2.45pt}
    \begin{subfigure}[b]{\mysubfloatwidthtwo}
        \fbox{\includegraphics[width=\textwidth]{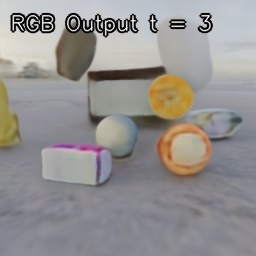}}
    \end{subfigure}
    \hfill 
    \begin{subfigure}[b]{\mysubfloatwidthtwo}
        \fbox{\includegraphics[width=\textwidth]{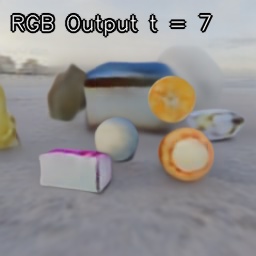}}
    \end{subfigure}
    \hfill 
    \begin{subfigure}[b]{\mysubfloatwidthtwo}
        \fbox{\includegraphics[width=\textwidth]{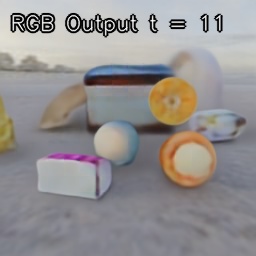}}
    \end{subfigure}
    \hfill 
    \begin{subfigure}[b]{\mysubfloatwidthtwo}
        \fbox{\includegraphics[width=\textwidth]{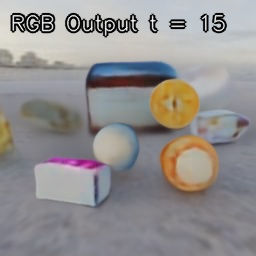}}
    \end{subfigure}
    \hfill 
    \begin{subfigure}[b]{\mysubfloatwidthtwo}
        \fbox{\includegraphics[width=\textwidth]{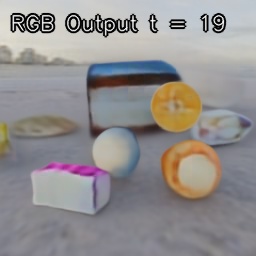}}
    \end{subfigure}
    \hfill 
    \begin{subfigure}[b]{\mysubfloatwidthtwo}
        \fbox{\includegraphics[width=\textwidth]{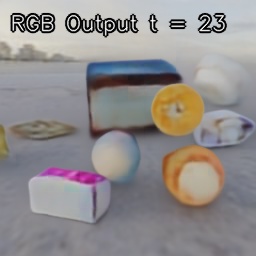}}
    \end{subfigure}
    \\
    \vspace{2.45pt}
    \begin{subfigure}[b]{\mysubfloatwidthtwo}
        \fbox{\includegraphics[width=\textwidth]{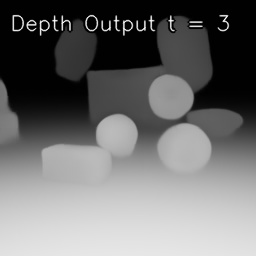}}
    \end{subfigure}
    \hfill 
    \begin{subfigure}[b]{\mysubfloatwidthtwo}
        \fbox{\includegraphics[width=\textwidth]{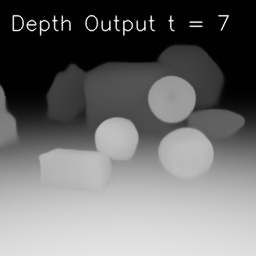}}
    \end{subfigure}
    \hfill 
    \begin{subfigure}[b]{\mysubfloatwidthtwo}
        \fbox{\includegraphics[width=\textwidth]{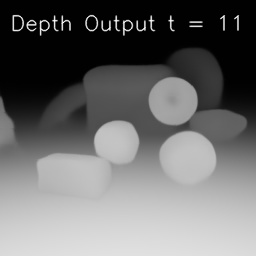}}
    \end{subfigure}
    \hfill 
    \begin{subfigure}[b]{\mysubfloatwidthtwo}
        \fbox{\includegraphics[width=\textwidth]{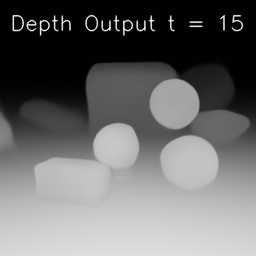}}
    \end{subfigure}
    \hfill 
    \begin{subfigure}[b]{\mysubfloatwidthtwo}
        \fbox{\includegraphics[width=\textwidth]{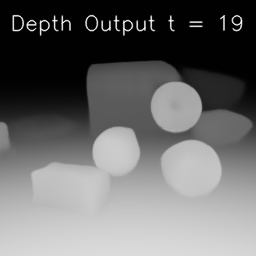}}
    \end{subfigure}
    \hfill 
    \begin{subfigure}[b]{\mysubfloatwidthtwo}
        \fbox{\includegraphics[width=\textwidth]{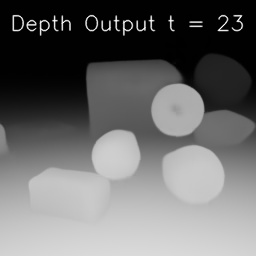}}
    \end{subfigure}
    \caption{Qualitative Analysis of Results on the MOVi-E Dataset: The top two rows show the input frames while the third row shows the target frame superimposed with the slot masks, and the last two rows show the next frame predictions.}
    \label{fig:movi_e_sequence}
\end{figure*}
\afterpage{\clearpage}

\section{Detailed \locis* Size and Wiring Information}
\label{app:lociswiring}
\begin{table*}
    \centering
    \caption{Encoder Architecture, for more information see the source file in \textit{nn/hyper\_encoder.py}}
    \label{tab:architecture}
    \begin{tabular}{lll}
    \toprule
    Component & Layer & Configuration \\
    \midrule
    \multirow{8}{*}{Encoder Base} 
        & ResidualPatchEmbedding & Conv2D(16, 32, 4, 4) + AvgPool + Channel Copy \\
        & HyperConvNext & 32 $\to$ 32 \\
        & ResidualPatchEmbedding & Conv2D(32, 64, 2, 2)  + AvgPool + Channel Copy \\
        & HyperConvNext & 64 $\to$ 64 \\
        & ResidualPatchEmbedding & Conv2D(64, 128, 2, 2)  + AvgPool + Channel Copy \\
        & HyperConvNext & 128 $\to$ 128 \\
        & HyperConvNext & 128 $\to$ 128 \\
        & HyperConvNext & 128 $\to$ 128 \\
    \midrule
    \multirow{4}{*}{Position Encoder}
        & HyperConvNext & 128 $\to$ 128 \\
        & HyperConvNext & 128 $\to$ 128 \\
        & HyperConvNext & 128 $\to$ 4 \\
        & FeaturesMapToPosition & \\
    \midrule
    \multirow{2}{*}{Gestalt Base Encoder}
        & HyperConvNext & 128 $\to$ 128 \\
        & HyperConvNext & 128 $\to$ 128 \\
    \midrule
    \multirow{6}{*}{Mask Gestalt Encoder}
        & ResidualPatchEmbedding & Conv2D(128, 256, 2, 2)  + AvgPool + Channel Copy \\
        & HyperConvNext & 256 $\to$ 256 \\
        & HyperConvNext & 256 $\to$ 256 \\
        & HyperConvNext & 256 $\to$ 256 \\
        & HyperConvNext & 256 $\to$ 256 \\
        & PositionPooling & \\
    \midrule
    \multirow{6}{*}{Depth Gestalt Encoder}
        & ResidualPatchEmbedding & Conv2D(128, 256, 2, 2)  + AvgPool + Channel Copy \\
        & HyperConvNext & 256 $\to$ 256 \\
        & HyperConvNext & 256 $\to$ 256 \\
        & HyperConvNext & 256 $\to$ 256 \\
        & HyperConvNext & 256 $\to$ 256 \\
        & PositionPooling & \\
    \midrule
    \multirow{6}{*}{RGB Gestalt Encoder}
        & ResidualPatchEmbedding & Conv2D(128, 256, 2, 2)  + AvgPool + Channel Copy \\
        & HyperConvNext & 256 $\to$ 256 \\
        & HyperConvNext & 256 $\to$ 256 \\
        & HyperConvNext & 256 $\to$ 256 \\
        & HyperConvNext & 256 $\to$ 256 \\
        & PositionPooling & \\
    \bottomrule
    \end{tabular}
\end{table*}

\begin{table*}
    \centering
    \caption{Predictor Architecture, for more information see the source file in \textit{nn/predictor.py}}
    \label{tab:additional_architecture}
    \begin{tabular}{lll}
    \toprule
    Component & Layer & Configuration \\
    \midrule
    \multirow{5}{*}{UpdateController} 
        & Linear & 1550 $\to$ 256 \\
        & SiLU & \\
        & Linear & 256  $\to$ 256 \\
        & SiLU & \\
        & Linear & 256  $\to$ 2 \\
    \midrule
    \multirow{16}{*}{Predictor}
        & InputEmbedding & \\
        & \quad Linear & 774  $\to$ 1024 \\
        & \quad SiLU & \\
        & \quad Linear & 1024  $\to$ 1024 \\
        & GateL0rd & 1024  $\to$ 1024 \\
        & MultiheadSelfAttention & 1024  $\to$ 1204 \\
        & GateL0rd & 1024  $\to$ 1024 \\
        & MultiheadSelfAttention & 1024 $\to$ 1204 \\
        & GateL0rd & 1024  $\to$ 1024 \\
        & MultiheadSelfAttention & 1024 $\to$ 1204 \\
        & GateL0rd & 1024 $\to$ 1024 \\
        & MultiheadSelfAttention & 1024 $\to$ 1204 \\
        & GateL0rd & 1024 $\to$ 1024 \\
        & OutputEmbedding & \\
        & \quad Linear & 1024 $\to$ 1024 \\
        & \quad SiLU & \\
        & \quad Linear & 1024 $\to$ 774 \\
    \bottomrule
    \end{tabular}
\end{table*}

\begin{table*}
    \centering
    \caption{Decoder Architecture, for more information see the source file in \textit{nn/decoder.py}}
    \label{tab:decoder_architecture}
    \begin{tabular}{lll}
    \toprule
    Component & Layer & Configuration \\
    \midrule
    \multirow{9}{*}{MaskDecoder} 
        & GestaltPositionFussion & \\
        & Conv2d & 256 $\to$ 128, kernel=3, pad=1 \\
        & SiLU & \\
        & Conv2d & 128 $\to$ 64, kernel=3, pad=1 \\
        & SiLU & \\
        & Conv2d & 64 $\to$ 32, kernel=3, pad=1 \\
        & SiLU & \\
        & Conv2d & 32 $\to$ 128, kernel=1 \\
        & TransposedConv2d & 128 $\to$ 1, kernel=16, stride=16 \\
    \midrule
    \multirow{15}{*}{DepthDecoder}
        & MaskEncoder & \\
        & \quad Conv2d & 1 $\to$ 128, kernel=16, stride=16 \\
        & \quad SiLU & \\
        & \quad Conv2d & 128 $\to$ 32, kernel=1 \\
        & GestaltMaskFussion & Gestalt * MaxPool(mask, kernel=16) \\
        & Concat & ModulatedGestalt, EncodedMask, PositionalEmbedding \\
        & Conv2d & 304 $\to$ 64, kernel=1 \\
        & ConvNeXt & 64 $\to$ 64 \\
        & ConvNeXt & 64 $\to$ 64 \\
        & ConvNeXt & 64 $\to$ 64 \\
        & Conv2d & 64 $\to$ 256, kernel=1 \\
        & SiLU & \\
        & TransposedConv2d & 256 $\to$ 1, kernel=16, stride=16 \\
    \midrule
    \multirow{19}{*}{RGBDecoder}
        & MaskEncoder & \\
        & \quad Conv2d & 1 $\to$ 128, kernel=16, stride=16 \\
        & \quad SiLU & \\
        & \quad Conv2d & 128 $\to$ 32, kernel=1 \\
        & DepthEncoder & \\
        & \quad Conv2d & 1 $\to$ 256, kernel=16, stride=16 \\
        & \quad SiLU & \\
        & \quad Conv2d & 256 $\to$ 64, kernel=1 \\
        & GestaltMaskFussion & Gestalt * MaxPool(mask, 16) \\
        & Concat & ModulatedGestalt, EncodedMask, EncodedDepth, PositionalEmbedding \\
        & Conv2d & 368 $\to$ 128, kernel=1 \\
        & ConvNeXt & 128 $\to$ 128 \\
        & ConvNeXt & 128 $\to$ 128 \\
        & ConvNeXt & 128 $\to$ 128 \\
        & ConvNeXt & 128 $\to$ 128 \\
        & ConvNeXt & 128 $\to$ 128 \\
        & Conv2d & 128 $\to$ 512, kernel=1 \\
        & SiLU & \\
        & TransposedConv2d & 512 $\to$ 3, kernel=16, stride=16 \\
    \bottomrule
    \end{tabular}
\end{table*}

\begin{table*}
    \centering
    \caption{Segmentation preprocessing, see \textit{nn/proposal\_v2.py} for more details}
    \label{tab:segmentation_unet_architecture}
    \begin{tabular}{lll}
    \toprule
    Component & Layer & Configuration \\
    \midrule
    \multirow{40}{*}{SegmentationUNet} 
        & Cat & Depth + 2D Grid(-1,1) \\
        & ResidualPatchEmbedding & Conv2D(3, 64, 4, 4) + AvgPool + Channel Copy \\
        & ConvNeXt & 64 $\to$ 64 \\
        & ResidualPatchEmbedding & Conv2D(64, 128, 2, 2) + AvgPool + Channel Copy \\
        & ConvNeXt & 128 $\to$ 128 \\
        & ConvNeXt & 128 $\to$ 128 \\
        & ResidualPatchEmbedding & Conv2D(128, 256, 2, 2) + AvgPool + Channel Copy \\
        & ConvNeXt & 256 $\to$ 256 \\
        & ConvNeXt & 256 $\to$ 256 \\
        & ConvNeXt & 256 $\to$ 256 \\
        & ResidualPatchEmbedding & Conv2D(256, 512, 2, 2) + AvgPool + Channel Copy \\
        & ConvNeXt & 256 $\to$ 256 \\
        & HyperNetwork & \\
        & \quad GlobalAvgPool & \\
        & \quad Linear & 512 $\to$ 512 \\
        & \quad SiLU & \\
        & \quad Linear & 512 $\to$ 512 \\
        & \quad SiLU & \\
        & \quad Linear & 512 $\to$ 512 \\
        & ConvNeXt & 512 $\to$ 512 \\
        & Conv2d & 512 $\to$ 2048, kernel=1 \\
        & SiLU & \\
        & Conv2d & 2048 $\to$ 256, kernel=2, stride=2 \\
        & ConcatFeatures & \\
        & Conv2d & 512 $\to$ 256, kernel=1 \\
        & ConvNeXt & 256 $\to$ 256 \\
        & Conv2d & 256 $\to$ 1024, kernel=1 \\
        & SiLU & \\
        & Conv2d & 1024 $\to$ 128, kernel=2, stride=2 \\
        & ConcatFeatures & \\
        & Conv2d & 256 $\to$ 128, kernel=1 \\
        & ConvNeXt & 128 $\to$ 128 \\
        & Conv2d & 128 $\to$ 512, kernel=1 \\
        & SiLU & \\
        & Conv2d & 512 $\to$ 64, 2, stride=2 \\
        & ConcatFeatures & \\
        & Conv2d & 128 $\to$ 64, kernel=1 \\
        & ConvNeXt & 64 $\to$ 64 \\
        & Conv2d & 64 $\to$ 512, kernel=1 \\
        & SiLU & \\
        & Conv2d & 512 $\to$ 32, kernel=4, stride=4 \\
        & ApplyHyperWeights & Features @ weights, 32, 16 \\
    \bottomrule
    \end{tabular}
\end{table*}

\begin{table*}
    \centering
    \caption{Background Architecture, for more information see the source file in \textit{nn/decoder.py}}
    \label{tab:bg_architecture}
    \begin{tabular}{lll}
    \toprule
    Component & Layer & Configuration \\
    \midrule
    \multirow{17}{*}{Uncertainty UNet} 
        & MaskEncoder & \\
        & ResidualPatchEmbedding & Conv2D(4, 16, 4, 4) + AvgPool + Channel Copy \\
        & ConvNeXt & 16 $\to$ 16 \\
        & ResidualPatchEmbedding & Conv2D(16, 32, 2, 2) + AvgPool + Channel Copy \\
        & ConvNeXt & 32 $\to$ 32 \\
        & ResidualPatchEmbedding & Conv2D(32, 64, 2, 2) + AvgPool + Channel Copy \\
        & ConvNeXt & 64 $\to$ 64 \\
        & ResidualPatchEmbedding & Conv2D(64, 128, 2, 2) + AvgPool + Channel Copy \\
        & ConvNeXt & 128 $\to$ 128 \\
        & ConvNeXt & 128 $\to$ 128 \\
        & ResidualPatchUpscaling & Conv2D(128, 64, 2, 2) + Upscale + Channel Avg \\
        & ConvNeXt & 64 $\to$ 64 \\
        & ResidualPatchUpscaling & Conv2D(64, 32, 2, 2) + Upscale + Channel Avg \\
        & ConvNeXt & 32 $\to$ 32 \\
        & ResidualPatchUpscaling & Conv2D(32, 16, 2, 2) + Upscale + Channel Avg \\
        & ConvNeXt & 16 $\to$ 16 \\
        & ResidualPatchUpscaling & Conv2D(16, 1, 4, 4) + Upscale + Channel Avg \\
    \midrule
    \multirow{6}{*}{\shortstack[c]{Background Extractor \\ Base-Encoder}}
        & PatchEmbedding & \\
        & \quad Conv2d & 4 $\to$ 256, kernel=16, stride=16 \\
        & \quad SiLU & \\
        & \quad Conv2d & 256 $\to$ 64, kernel=1 \\
        & MHA-Layer & 64 $\to$ 64 \\
        & MHA-Layer & 64 $\to$ 64 \\
    \midrule
    \multirow{2}{*}{\shortstack[c]{Background Extractor \\ RGB-Encoder}}
        & MHA-Layer & 64 $\to$ 64 \\
        & & \\
    \midrule
    \multirow{4}{*}{\shortstack[c]{Background Extractor \\ Depth-Encoder}}
        & MHA-Layer & 64 $\to$ 64 \\
        & Bottleneck & token avg + cross attention to single token \\
        & Sigmoid &  \\
        & Binarize & $    x \gets x + x (1 - x) \mathcal{N}(0, 1)$ \\
    \midrule
    \multirow{6}{*}{\shortstack[c]{Background Extractor \\ Depth-Decoder}}
        & ConvNeXt & 64 $\to$ 64 \\
        & ConvNeXt & 64 $\to$ 64 \\
        & PatchUpscaling & \\
        & \quad Conv2d & 64 $\to$ 256, kernel=1 \\
        & \quad SiLU & \\
        & \quad TransposedConv2d & 256 $\to$ 1, kernel=16, stride=16 \\
    \midrule
    \multirow{10}{*}{\shortstack[c]{Background Extractor \\ RGB-Decoder}}
        & Depth-Encoder & \\
        & \quad PatchEmbedding & 1 $\to$ 64 \\
        & \quad ConvNeXt & 64 $\to$ 64 \\
        & Cross-Attention-Layer & 64 $\to$ 64 \\
        & ConvNeXt & 64 $\to$ 64 \\
        & ConvNeXt & 64 $\to$ 64 \\
        & PatchUpscaling & \\
        & \quad Conv2d & 64 $\to$ 256, kernel=1 \\
        & \quad SiLU & \\
        & \quad TransposedConv2d & 256 $\to$ 1, kernel=16, stride=16 \\
    \bottomrule
    \end{tabular}
\end{table*}

\end{document}